\documentclass[11pt,a4paper]{article}
\usepackage[margin=1.1in]{geometry}
\usepackage[utf8]{inputenc}
\usepackage[T1]{fontenc}
\usepackage{mathpazo}
\usepackage{hyperref}
\usepackage{url}
\usepackage{amsfonts,amssymb,amsthm,amsmath}
\usepackage{nicefrac}
\usepackage{microtype}
\usepackage{natbib}
\usepackage[dvipsnames]{xcolor}
\usepackage{graphicx}
\usepackage{caption}
\usepackage[position=top]{subfig}
\usepackage{bm}
\usepackage[inline]{enumitem}
\usepackage{algorithm}
\usepackage[noend]{algpseudocode}
\usepackage{tikz}
\usetikzlibrary{decorations.pathmorphing,decorations.shapes}

\algnewcommand\algorithmicinput{\textbf{Input:}}
\algnewcommand\Input{\item[\algorithmicinput]}

\algnewcommand\algorithmicinputx{\textbf{Input}}
\algnewcommand\Inputx{\item[\algorithmicinputx]}

\usepackage{xpatch}
\makeatletter
\xpatchcmd{\algorithmic}{\itemsep\z@}{\itemsep=0.7ex plus1pt}{}{}
\makeatother

\newcommand{\bN}{\mathbb{N}}
\newcommand{\bR}{\mathbb{R}}
\DeclareMathOperator{\E}{\mathbb{E}}
\DeclareMathOperator{\given}{ | }
\DeclareMathOperator{\dP}{P}
\DeclareMathOperator{\prob}{P}
\newcommand{\usub}[2]{\underset{#1}{\underbrace{#2}}}
\newcommand{\B}[1]{\bm{#1}}
\newcommand{\btheta}{\ensuremath \B{\theta}}
\newcommand{\bphi}{\ensuremath \B{\phi}}
\newcommand{\zspace}{\ensuremath \mathcal{Z}}
\newcommand{\xspace}{\ensuremath \mathcal{X}}
\newcommand{\yspace}{\ensuremath \mathcal{Y}}
\newcommand{\cD}{\ensuremath \mathcal{D}}
\newcommand{\pitil}{\ensuremath \tilde{\pi}}
\newcommand{\Qcal}{\mathcal{Q}}
\newcommand{\xhdr}[1]{\noindent{\textbf{#1.}}}
\newcommand{\secref}[1]{Section~\ref{#1}}
\def\eqp{\: .}
\def\eqc{\: ,}

\newtheorem{proposition}{Proposition}
\newtheorem{corollary}[proposition]{Corollary}
\newtheorem*{corollary*}{Corollary}
\newtheorem*{proposition*}{Proposition}

\graphicspath{{figs/}}

\title{\huge Fair Decisions Despite Imperfect Predictions}

\author{\large
Niki Kilbertus\textsuperscript{$\dagger\ddagger$} \hspace{0.5cm}
Manuel Gomez-Rodriguez\textsuperscript{$\sharp$} \\
\large
Bernhard Sch\"{o}lkopf\textsuperscript{$\dagger$} \hspace{0.5cm}
Krikamol Muandet\textsuperscript{$\dagger$} \hspace{0.5cm}
Isabel Valera\textsuperscript{$\dagger$} \hspace{0.5cm} \\[0.5cm]
\normalsize\textsuperscript{$\dagger$}Max Planck Institute for Intelligent Systems, T\"ubingen, Germany\\
\normalsize\textsuperscript{$\ddagger$}Department of Engineering, University of Cambridge, United Kingdom\\
\normalsize\textsuperscript{$\sharp$}Max Planck Institute for Software Systems, Kaiserslautern, Germany\\[-0.3cm]
}

\date{}

\begin{document}

\maketitle

\begin{abstract}
	\noindent Consequential decisions are increasingly informed by sophisticated data-driven predictive models.
	However, to consistently learn accurate predictive models, one needs access to ground truth labels.
	Unfortunately, in practice, labels may only exist conditional on certain decisions---if a loan is denied, there is not even an option for the individual to pay back the loan.
	Hence, the observed data distribution depends on how decisions are being made.
	In this paper, we show that in this \emph{selective labels} setting, learning a predictor directly only from available labeled data is suboptimal in terms of both fairness and utility.
	To avoid this undesirable behavior, we propose to directly learn decision policies that maximize utility under fairness constraints and thereby take into account how decisions affect which data is observed in the future.
	Our results suggest the need for a paradigm shift in the context of fair machine learning from the currently prevalent idea of simply building predictive models from a single static dataset via risk minimization, to a more interactive notion of ``\emph{learning to decide}''.
	In particular, such policies should not entirely neglect part of the input space, drawing connections to explore/exploit tradeoffs in reinforcement learning, data missingness, and potential outcomes in causal inference.
	Experiments on synthetic and real-world data illustrate the favorable properties of learning to decide in terms of utility and fairness.
\end{abstract}

\section{Introduction}
\label{sec:intro}

The use of machine learning models to assist consequential decision making---where decisions have significant consequences for individuals---is becoming common in a variety of critical applications.
We start by revisiting some common settings of consequential decisions that may be (partially) automated.
In pretrial release decisions, a judge may consult a learned model of the probability of recidivism to decide whether to grant bail or not.
In loan decisions, a bank may decide whether or not to offer a loan based on learned estimates of the credit default probability.
In fraud detection, an insurance company may flag suspicious claims based on a machine learning model'{}s predicted probability that the claim is fraudulent.
In all these scenarios, the goal of the decision maker (bank, law court, or insurance company) may be to take decisions that maximize a given utility function.
Since such a utility is a function of the entire policy, rather than an individual loss term for each prediction in isolation, it may encode preferences that go beyond merely ``correct or incorrect'' for a predictive classification task.
For example, the utility may accommodate various fairness and diversity considerations, or enforce actions that could improve the wellbeing of certain individuals and groups in the long run---even when such actions are not warranted by the directly observed outcomes.
In contrast, the goal of a supervised predictive machine learning model is solely to provide accurate predictions given the available training set, typically under the implicit assumption that the training data is an i.i.d.\ sample from the distribution encountered during test time.
Such a simplistic approach, which we refer to as \emph{learning to predict}, ignores that once decisions are based on these predictions, they may interact with the data collection or have direct impact on the underlying distribution relevant during deployment \citep{perdomo2020performative}.

In this context, there has been much work on computational mechanisms to ensure that machine learning models do not disproportionately harm particular demographic groups sharing one or more sensitive attributes, e.g., race or gender \citep{Dwork2012,Feldman2015}.
Nevertheless, most work on fair machine learning does not distinguish between decisions and label predictions, which also leads to a perceived trade-off between fairness and accuracy or performance \citep{Chouldechova2017,Kleinberg2016}.
This stems from the fact that viewing ``incorrect'' predictions as positively desirable, e.g., because they may promote fairness, is incompatible with the standard goal of minimizing a predictive loss in supervised learning.
From that viewpoint, an accurate prediction is equivalent---or at least directly translates---to a good decision.
Only recently has the distinction been made explicit, typically emphasizing that \emph{not predicting historically recorded labels correctly} can often be part of the goal when it comes to fairness \citep{corbett2017algorithmic,kleinberg2017human,mitchell2018prediction,valera2018enhancing}.
This recent line of work has shown that if a predictive model achieves perfect prediction accuracy, \emph{deterministic threshold rules}, which derive decisions deterministically from the predictive model by thresholding, indeed achieve maximum utility under various fairness constraints.
At first, this lends support to focusing on deterministic threshold rules and seemingly justifies using predictions and decisions interchangeably.

However, in many practical scenarios including the ones described in the beginning of this section, the decision determines whether a label is realized or not---if bail (a loan) is denied, there is not even an option for the individual to reoffend (pay back the loan).
This problem has been referred to by \citet{lakkaraju2017selective} as \emph{selective labels}.
As a consequence, the labeled data used to train predictive models often depend on the decisions taken, which likely leads to suboptimal performance.
For example, a racist initial policy may categorically reject applicants from a certain demographic group.
Therefore, no data about these individuals is collected and there are no guarantees for how a predictive model may extrapolate into this unseen region of the input space.
Indeed, deterministic threshold rules using even slightly imperfect predictive models can be far from optimal \citep{woodworth2017learning}.
This negative result raises the following question: \textit{Can we do better if we learn directly to decide rather than to predict?}\footnote{We remark that our notion of learning to decide is not immediately related to (Bayesian) decision theory. Moreover, for this paper, we restrict the notion of learning to predict to simple point estimates (categorical predictions) from risk minimization based on available data. Crafting ``predictions'' more carefully, for example by regarding a data-missingness model or proper uncertainty estimates, may not be prone to the same issues. However, most works on fairness in machine learning have been staged within our simplistic ``learning to predict'' framework.}
Here, by ``learning to decide'' we mean learning to maximize a utility of a decision policy (rather than predictions) when deployed under test time conditions (rather than merely trained on observed data).

In the present work, we first articulate how the ``learning to predict'' approach fails in a utility maximization setting (with fairness constraints) that accommodates a variety of real-world applications, including those mentioned previously.
We show that label data gathered under deterministic rules (e.g., prediction based threshold rules) are neither sufficient to improve the accuracy of the underlying predictive model, nor the utility of the decision making process.
We then demonstrate how to overcome this undesirable behavior using a particular family of stochastic decision rules and introduce a simple gradient-based algorithm to learn them from data.
Experiments on synthetic and real-world data illustrate our theoretical results and show that, under imperfect predictions, \emph{learning to predict} is inferior to \emph{learning to decide}.
Code is available at \url{https://github.com/nikikilbertus/fair-decisions}.

\xhdr{Related work}
The work most closely related to ours analyzes the long-term effects of consequential decisions informed by data-driven predictive models on underrepresented groups \citep{hucheng2018,liu18c,mouzannar2019fair}.
However, this line of work focuses mainly on the evolution of several measures of well-being under a perfect predictive model and neglecting the data collection phase.
In contrast, we focus on analyzing how to improve a suboptimal decision process when labels exist only for positive decisions.
Potential issues arising from neglecting the data collection process have also been highlighted in a survey of what machine learning practitioners in industry actually need to enforce fairness \citep{holstein2018improving}.
\citet{dimitrakakis2019bayesian} similarly point out how attempts of fair machine learning may fail when there is uncertainty about the underlying probabilistic model of the world.
They achieve fairness in such settings by deploying a Bayesian approach that aims at enforcing fairness in all possible models weighted by their probability given the current information.
In contrast, we focus on analyzing how to improve a suboptimal decision process when labels exist only for positive decisions.
More broadly, our work relates to the growing literature on fairness in machine learning, which mostly attempts to match various statistics of the predictive models across protected subgroups.

We also build on previous work on counterfactual inference and policy learning \citep{athey2017efficient,ensign2017decision,gillen2018online,heidari2018preventing,joseph2016fairness,jung2018algorithmic,kallus2018balanced,kallus2018residual,lakkaraju2017learning}.
In these settings, the decision typically determines which of the potential outcomes is observed and the focus is on confounders that affect both the decision and the outcome \citep{Rubin05:PO}.
In contrast, in our approach the decision determines whether there will be an outcome at all, but there is no unobserved confounding.
Two notable exceptions are by \citet{kallus2018residual} and \citet{ensign2017decision}, which also consider limited feedback.
However, \citet{kallus2018residual} focus on designing unbiased estimates for fairness measures, rather than learning how to decide.
\citet{ensign2017decision} assume a deterministic mapping between features and labels, which allows them to reduce the problem to the apple tasting problem \citep{helmbold2000apple}.
Remarkably, in their deterministic setting, they also conclude that the optimal decisions should be stochastic.

Unlike in the fairness literature, where deterministic policies dominate \citep{corbett2017algorithmic,valera2018enhancing,Meyer2018objecting}, stochastic policies are often necessary to ensure adequate exploration \citep{Silver14:DPG} in contextual bandits \citep{Dudik11:DoublyRobust,Langford08:ES,Agarwalb14:Monster} and reinforcement learning \citep{jabbari2016fairness,Sutton98:RL}.
While the techniques applied there provide valuable pointers, the typical problem setting differs fundamentally from ours and usually neither fairness constraints nor selective labels are taken into account.
A recent notable exception is \citet{joseph2016fairness}, initiating the study of fairness in multi-armed bandits, however, using a fairness notion orthogonal to the observational group matching criteria we consider in our work, and ignoring the selective labels problem.

The crucial difference between mere predictions and actual decisions has been further highlighted by, e.g., \citet{kleinberg2018algorithmic,rambachan2020economic}.
There, the authors argue for a social planner taking decisions to maximize a social welfare function that may also include fairness preferences.
If the social planner has access to predictions from machine learning systems, it is optimal to keep the decision and prediction stage completely separate and train the machine learning pipeline to maximize predictive performance without any additional constraints.
This suggests that machine learning systems should indeed be used for prediction alone, but there is a separate optimization problem in deriving decisions from these predictions.
It is this second stage in which we must account for fairness considerations.
However, these approaches also do not consider selective labels and the created need for exploration.
In general, as soon as the observed data depends on the chosen decision rule, the prediction and decision stage may not be easily decoupled anymore.
Merely adjusting or improving predictive models does not suffice to make beneficial decisions.

\section{Decisions from imperfect predictive models}
\label{sec:formulation}

Let $\xspace \subseteq \bR^d$ be the feature domain, $\zspace = \{0,1\}$ the range of sensitive attributes,\footnote{For simplicity, we assume the sensitive attribute to be binary, potentially resulting in inadequate binary gender or race assignments.
Our work can easily be extended to categorical sensitive attributes.} and $\yspace = \{0,1\}$ the set of ground truth labels.
We assume the standard sigma algebras on these spaces.
A \emph{decision rule} or \emph{policy}\footnote{We use the terms \emph{decision rule}, \emph{decision making process} and \emph{policy} interchangeably in this paper.}
is a mapping $\pi: \xspace \times \zspace \to \mathcal{P}(\{0,1\})$ that maps an individual'{}s feature vector and sensitive attribute to a probability distribution over \emph{decisions} $d \in \{0,1\}$.
For each individual, we sample $x, z$ and $y$ from a ground truth distribution $\dP(X,Z,Y) = \dP(Y\given X,Z) \dP(X,Z)$.
Decisions $d$ are sampled from a policy $d \sim \pi(D\given x, z)$, where we often write $\pi(x,z)$ for $\pi(D\given x, z)$ and $\pi(D=1 \given x, z)$ for the probability of a positive decision given features~$x, z$.
The decision determines whether the label $y \sim \dP(Y\given X,Z)$ comes into existence.
In loan decisions, the feature vector $x$ may include salary, education, or credit history;
the sensitive attribute $z$ may indicate sex;
a loan can be granted ($d = 1$) or denied ($d = 0$);
and the label $y$ indicates repayment ($y = 1$) or default ($y=0$) \emph{upon receiving a loan}.

Inspired by \citet{corbett2017algorithmic}, we measure the \emph{utility} as the expected overall profit provided by the policy with respect to the distribution $\dP$, i.e.,
\begin{equation}\label{eq:utility}
  u_{\dP}(\pi) := \E_{x,z,y \sim \dP,\, d\sim \pi(x, z)} \left[y\, d - c\, d \right]
  = \E_{x,z \sim \dP} \left[\pi(D = 1 \given x, z) (\prob(Y = 1 \given x, z) - c) \right] \eqc
\end{equation}
where $c \in (0, 1)$ reflects economic considerations of the decision maker.
For example, in a loan scenario, the utility gain is $(1-c)$ if a loan is granted and repaid, $-c$ if a loan is granted but the individual defaults, and zero if the loan is not granted.
One could think of adding a term for negative decisions of the form $g(y) (1-d)$ for some given definition of $g$, however, we would not be able to compute such a term due to the selective labels, except for constant $g$.
Therefore, without loss of generality, we assume that $g(y)= 0$ for all $y$, because any non-zero constant $g$ can easily be absorbed in our framework.

For fairness considerations, we define the \emph{$f$-benefit for group $z \in \{0,1\}$} with respect to the distribution $\dP$ by
\begin{equation*}
  b_{\dP}^z(\pi) := \E_{x, y \sim \dP(X, Y \given z),\, d\sim \pi(x, z)} [f(d, y)] \eqc
\end{equation*}
with $f: \{0, 1\} \times \{0, 1\} \to \bR$.
Note that common observational group matching fairness criteria can be expressed as $b_{\dP}^0(\pi) = b_{\dP}^1(\pi)$ for different choices of $f$ (and perhaps conditioning on specific values of $y$ or $d$ corresponding to criteria of separability or sufficiency respectively \citep{barocas-hardt-narayanan}).
For simplicity, we will focus on demographic parity (or no disparate impact), which simply amounts to $f(d, y) = d$.

Under perfect knowledge of $\dP(Y \given x, z)$, the policy maximizing the above utility subject to the group benefit fairness constraint $b_{\dP}^0(\pi) = b_{\dP}^1(\pi)$ is a deterministic threshold rule \citep{corbett2017algorithmic}\footnote{Here, $\B{1}[\bullet]$ is $1$ if the predicate $\bullet$ is true and $0$ otherwise.}
\begin{equation}\label{eq:detthresh}
  \pi^*(D = 1 \given x, z) = \B{1}[\prob(Y = 1 \given x, z) \geq c_z] \eqc
\end{equation}
where we allow for group specific cost factors $c_0, c_1$ such that $b_{\dP}^0(\pi) = b_{\dP}^1(\pi)$.
Without fairness constraints, we simply have $c_0 = c_1 = c$.
However, as discussed by \citet{woodworth2017learning}, in practice we typically do not have access to the true conditional distribution $\dP(Y \given x, z)$, but instead to an imperfect predictive model $Q(Y \given x, z)$ trained on a finite training set.
Such a predictive model can similarly be used to implement a deterministic threshold rule as
\begin{equation}\label{eq:detthreshQ}
  {\pi}_{Q}(D = 1 \given x, z) = \B{1}[Q(Y = 1 \given x, z) \geq c] \eqp
\end{equation}
Here, the predictor $Q(Y = 1 \given x, z) \approx \dP(Y = 1 \given x, z) - \delta_z$, with $\delta_z = c_z - c$, directly incorporates the fairness constraint, i.e., it is trained to maximize predictive power subject to the fairness constraint.
In this context, \citet{woodworth2017learning} have shown that this approach often leads to better performance than post-processing a potentially unfair predictor as proposed by \citet{Hardt2016}.
Unfortunately, they have also shown that, because of the mismatch between $Q(Y = 1 \given x, z)$ and $\dP(Y = 1 \given x, z) - \delta_z$, the resulting policy $\pi_Q$ will usually still be suboptimal in terms of both utility and fairness.
To make things worse, due to the selective labeling, the data points $x, z, y$ observed under a given policy $\pi_0$ are not i.i.d.\ samples from the ground truth distribution $\dP(X, Z, Y)$, but instead from the weighted distribution
\begin{equation}\label{eq:imperfect-p}
  \dP_{\pi_0}(X,Z,Y) \propto \dP(Y \given X,Z)\, {\pi_0}(D=1\given X,Z)\, \dP(X,Z) \eqp
\end{equation}
Consequently, if $\pi_0$ is not optimal, i.e., $\pi_0 \ne \pi^*$, the necessary i.i.d.\ assumption for consistency results of empirical risk minimization is violated, which may also be one reason for a common observation in fairness, namely that predictive errors are often systematically larger for minority groups \citep{Angwin2016}.
In the remainder, we will say that the distributions $\dP_{\pi_0}(X,Z,Y)$ and $\dP_{\pi_0}(X,Z)$ are \emph{induced} by the policy $\pi_0$.
In the next section, we study how to learn the optimal policy, potentially subject to fairness constraints, if the data is collected from an initial faulty policy $\pi_0$.

\section{From deterministic to stochastic policies}
\label{sec:sequential}

Consider a class of policies $\Pi$, within which we want to maximize utility, as defined in eq.~\eqref{eq:utility} subject to the group benefit fairness constraint $b_{\dP}^0(\pi) = b_{\dP}^1(\pi)$.
We formulate this as an unconstrained optimization with an additional penalty term, namely to maximize
\begin{equation}\label{eq:policy_learning}
  v_{\dP}(\pi) := u_{\dP}(\pi) - \frac{\lambda}{2} (b_{\dP}^0(\pi) - b_{\dP}^1(\pi))^2
\end{equation}
over $\pi \in \Pi$ under the assumption that we do not have access to samples from the ground truth distribution $\dP(X, Z, Y)$, which $u_{\dP}(\pi)$ and $b_{\dP}^z(\pi)$ depend on.
Instead, we only have access to samples from a distribution $\dP_{\pi_0}(X, Z, Y)$ induced by a given initial policy $\pi_0$ as in eq.~\eqref{eq:imperfect-p}.
We first analyze this problem for deterministic threshold rules, before considering general deterministic policies, and finally also general stochastic policies.

\subsection{Deterministic policies}

Assume the initial policy $\pi_0$ is a given deterministic threshold rule and $\Pi$ is the set of all deterministic threshold rules, which means that each $\pi \in \Pi$ (and $\pi_0$) is of the form eq.~\eqref{eq:detthreshQ} for some predictive model $Q(Y \given x, z)$.
Given a hypothesis class of predictive models $\Qcal$, we reformulate eq.~\eqref{eq:policy_learning} to maximize
\begin{equation}\label{eq:policy_learning-p}
  v_{\dP}(\pi_Q) := u_{\dP}(\pi_Q) - \frac{\lambda}{2} (b_{\dP}^0(\pi_Q) - b_{\dP}^1(\pi_Q))^2
\end{equation}
over $Q \in \Qcal$, where the utility and the benefits for $z\in \{0,1\}$ are simply $u_{\dP}(\pi_Q) = \E_{x, z, y \sim \dP} [ \B{1}[Q(Y = 1\given x, z) \ge c] (y - c)]$ and $b_{\dP}^z(\pi_Q) = \E_{x, z, y \sim \dP} [f(\B{1}[Q(Y = 1\given x, z) \ge c], y)]$.
Note that eq.~\eqref{eq:policy_learning} has a unique optimum $\pi^*$ (up to differences on sets of measure zero).
Therefore, if $\pi^* \in \Pi$ (the set of all deterministic threshold rules), eq.~\eqref{eq:policy_learning-p} will also reach this optimum if $\Qcal$ is rich enough.
However, the optimal predictor $Q^*$ may not be unique, because the utility and the benefits are not sensitive to the precise values of $Q(Y = 1 \given x, z)$ above or below the threshold $c$.
We may have that a $Q \in \Qcal$ with $Q \ne Q^*$ may approximate $\dP(Y \given x, z) - \delta_z$ more accurately than $Q^{*}$.

If we only have access to samples from the distribution $\dP_{\pi_0}$ induced by some $\pi_0 \ne \pi^*$, we may choose to simply learn a predictive model $Q_0^* \in \Qcal$ that empirically maximizes the objective $v_{\dP_{\pi_0}}(\pi_Q)$, where the utility and the benefits are computed with respect to the induced distribution $\dP_{\pi_0}$.
However, the following negative result shows that, under mild conditions, $Q^*_0$ leads to a suboptimal deterministic threshold rule.
\begin{proposition}\label{prop:limitation}
If there exists a subset $\mathcal{V} \subset \xspace \times \zspace$ of positive measure under $\dP$ such that $\dP(Y=1 \given \mathcal{V}) \ge c$ and $\dP_{\pi_0} (Y = 1\given \mathcal{V}) < c$, then there exists a maximum $Q_0^* \in \mathcal{Q}$ of $v_{\dP_{\pi_0}}$ such that $v_{\dP}(\pi_{Q_0^*}) < v_{\dP}(\pi_{Q^*})$.
\end{proposition}
\begin{proof}
First, note that any deterministic policy $\pi$ is fully characterized (up to differences of measure zero) by the sets $W_d(\pi) = \{(x,z) \given \pi(D = 1 \given x,z) = d\}$ for $d \in \{0, 1\}$.
For a deterministic threshold rule $\pi_Q$, we write $W_d(Q) = \{(x,z) \given \B{1}[Q(Y = 1\given x,z) > c] = d\} = W_d(\pi_Q)$.
By definition, we have that $v(\pi_{Q}) \le v(\pi_{Q^*})$.
We note that whenever the symmetric difference between the sets $W_d(Q)$ and $W_d(Q^*$), $W_d(Q) \Delta W_d(Q^*)$, has positive inner measure (induced by $\dP$) for $d \in \{0,1\}$ and a $Q \in \Qcal$, we have $v(\pi_{Q}) \ne v(\pi_{Q^*})$ and thus $v(\pi_{Q}) < v(\pi_{Q^*})$.
Thus it only remains to show that $W_d(Q^*) \Delta W_d(Q_0^*)$ has positive inner measure for $d \in \{0,1\}$.
Since $\dP(Y = 1 \given \mathcal{V}) \ge c$ by assumption, we have $\mathcal{V} \subset W_1(Q^*)$.
At the same time, because of $\dP_{\pi_0}(Y = 1 \given \mathcal{V}) < c$ by assumption, we have $\mathcal{V} \cap W_1(\pi_0) = \emptyset$.
Finally, we note that for any $Q\in \mathcal{Q}$, we have that $v_{\dP_{\pi_0}}(Q) = v_{\dP_{\pi_0}}(Q \cdot \chi_{W_1(\pi_0)})$, where $\chi_{\bullet}$ is the indicator function on the set $\bullet$.
Therefore, we can choose a $Q^*_0$ maximizing $v_{\dP_{\pi_0}}$ such that $W_1(Q^*_0) \subset W_1(\pi_0)$ and thus $\mathcal{V} \cap W_1(Q^*_0) = \emptyset$.
Therefore $\mathcal{V} \subset W_1(Q_0^*) \Delta W_1(Q^*)$ and $\mathcal{V}$ has positive measure under $\dP$ by assumption.
Thus $W_d(Q_0^*) \Delta W_d(Q^*)$ has positive inner measure and we conclude $v_{\dP}(\pi_{Q_0^*}) < v_{\dP}(\pi_{Q^*})$.
\end{proof}

\begin{figure}
\centering
\includegraphics[width=0.32\columnwidth]{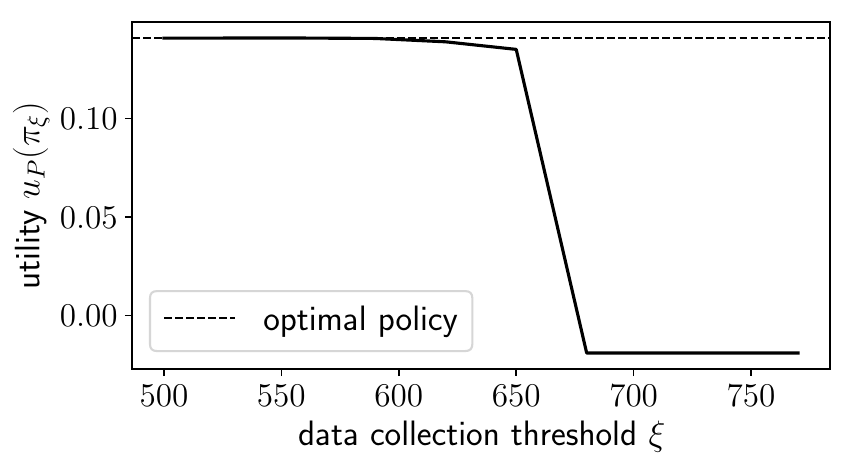}%
\includegraphics[width=0.32\columnwidth]{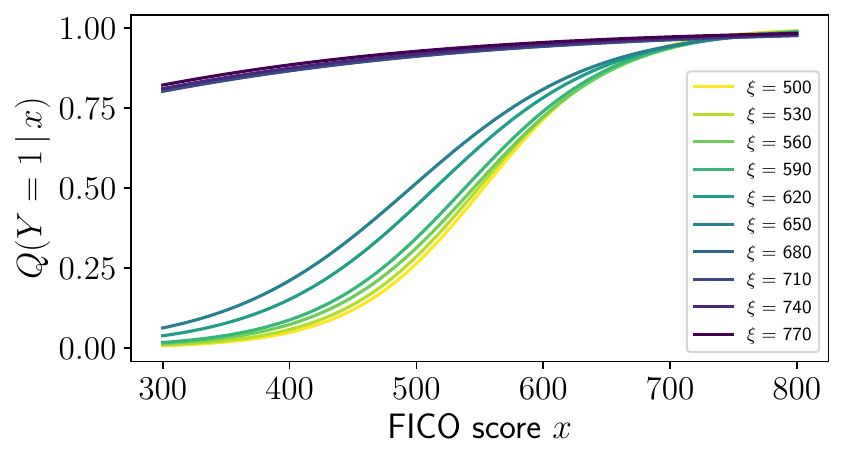}%
\includegraphics[width=0.32\columnwidth]{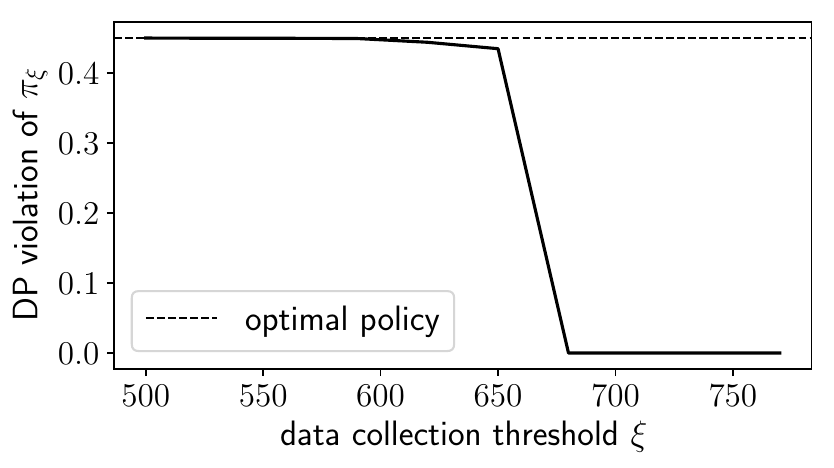}%
\caption{We show the utility (left) and the predictive models $Q_{\xi}$ learned from data collected with an initial threshold of $\xi$ (middle).
Finally, we present the violation of demographic parity (right) of threshold decision rules $\pi_{\xi}$ learned from data collected with an initial threshold of $\xi$.
Harsh data collection policies (i.e., large $\xi$)---while achieving
demographic parity---render the learned policies useless in terms of utility.
}
\label{fig:fico-example}
\end{figure}

\textbf{Lending example.} We briefly illustrate this result in a lending example based on FICO credit score data as described in \citet{Hardt2016}.
Such single feature scenarios are highly relevant for score-based decision support systems where full training data and the functional form of the score are often not available (e.g., also for pretrial risk assessment).
For any score that is strictly monotonic in the true success rate, the optimal policy is simply to threshold the score.
This lends additional support to score-based systems.

Here, we can generate new scores for a given group via inverse transform sampling from the known cumulative distribution functions.
We consider $80$\% white and $20$\% black applicants.
A hypothetical new bank that has access to FICO scores $x \in \xspace := \{300, \ldots, 820\}$, but not to the corresponding repayment probabilities may expect to be profitable if at least $70$\% of granted loans are repaid, i.e., $c = 0.7$.
A risk-averse lender may initially choose a high score threshold $\xi \in \xspace$ and employ the decision rule $\B{1}[x > \xi]$.
After collecting repayment data $\cD^{(\xi)} := \{(x_i, y_i)\}_{i = 1}^n$ with this initial threshold, they learn a model $Q_{\xi}(Y = 1 \given x)$ and then decide based on $\pi_{\xi}(D = 1 \given x) = \B{1}[Q_{\xi}(Y = 1 \given x) > c]$.

For a range of initial data collection score thresholds $\xi \in [500, 800]$, we sample 10,000 scores from the specified population ($80$\% white, $20$\% black) via inverse transform sampling given the cumulative distributions functions over scores of the two groups.
The relatively large number of examples is chosen to illustrate that the negative result is not a consequence of insufficient data.
We then fit an L2 regularized logistic regression model to each of these datasets using 5-fold cross validation to select the regularization parameter.
This results in a predictive model $Q_{\xi}$ for each initial data collection threshold $\xi$.
For each of these models we construct the decision rule $\pi_{\xi}(D = 1 \given x) = \B{1}[Q_{\xi}(Y = 1 \given x) > c]$, with $c=0.7$.
We then estimate utility and fairness violation of demographic parity on a large sample from the entire population (one million examples).

In Figure~\ref{fig:fico-example} we show how the initial data collection threshold $\xi$ affects utility and fairness of the resulting predictive model-based decision rule.
Conservatively high initial thresholds of $\xi \ge 650$ lead to essentially useless decisions $\pi_{\xi}$, because of imperfect prediction models regardless of how much data was collected.
More lenient initial policies can result in near optimal decisions with improved fairness compared to the maximum utility policy for the given cost $c$ (dashed).

A simple fix seems to present itself: Do not start with high thresholds.
However, Proposition~\ref{prop:limitation} tells us that it does not matter how low we set the initial threshold, if we use it to derive deterministic decisions.
By deterministically thresholding the score, we inevitable reject a subset of the population (with positive measure) and thus can never learn whether some of them may actually repay.
This will be highlighted in the next paragraphs, showing practical impossibility results for recovering from a bad initial policy even in a sequential training setting when using deterministic decision rules.
The only way to overcome this issue in our example is not to draw a hard threshold, bat to accept applicants with some non-zero probability for \emph{every possible score}.
We will formalize this idea and its advantages in \secref{sec:exploring}.

\textbf{Impossibility results.} Supplementing the result in Proposition~\ref{prop:limitation}, we will now prove that---in certain situations---a sequence of deterministic threshold rules, fails to recover the optimal policy despite it being in the hypothesis class.
We assume that each threshold rule is of the form of eq.~\eqref{eq:detthreshQ} and its associated predictive model is trained using the data gathered through the deployment of previous threshold rules.
To this end, we consider a \emph{sequential policy learning task}, which is given by a tuple $(\pi_0, \Pi', \mathcal{A})$, where:
\begin{enumerate}[label=\alph*)]
  \item $\Pi' \subset \Pi$ is the hypothesis class of policies,
  \item $\pi_0 \in \Pi'$ is the initial policy, and
  \item $\mathcal{A}: \Pi' \times \bigcup_{i=1}^{\infty} (\xspace \times \zspace \times \yspace)^i \to \Pi'$ is an update rule.
\end{enumerate}
The update rule $\mathcal{A}$ takes an existing policy $\pi_t$ and a dataset $\cD \in (\xspace \times \zspace \times \yspace)^n$ and produces an updated policy $\pi_{t+1}$, which typically aims to improve the policy in terms of the objective function $v_{\dP}(\pi)$ in eq.~\eqref{eq:policy_learning}.
In our setting, the dataset $\cD$ is collected by deploying previous policies, i.e., from a mixture of the distributions $\dP_{\pi_{\tau}}(X, Z, Y)$ with $\tau \le t$.

Recall that for deterministic threshold policies we can partition the space $\xspace \times \zspace = W_0(\pi)\cup W_1(\pi)$ into regions of negative and positive decisions.
Then, we say an update rule is \emph{non-exploring on $\cD$} if and only if $W_0(\mathcal{A}(\pi, \cD)) \subset W_0(\pi)$.
Intuitively, this means that no individual who has received a negative decision under the old policy $\pi$ would receive a positive decision under the new policy $\mathcal{A}(\pi, \cD)$.
Common learning algorithms for classification, such as gradient boosted trees are \emph{error based}, i.e., they only change the decision function when they make errors on the training set.
As a result, they lead to non-exploring update rules on $\cD$ whenever they achieve zero error.
\begin{proposition}\label{prop:limits}
Let $(\pi_0, \Pi', \mathcal{A})$ be a sequential policy learning task, where $\Pi' \subset \Pi$ are deterministic threshold policies based on a class of predictive models, and let the initial policy be more strict than the optimal one, i.e., $W_0(\pi_0) \supsetneq W_0(\pi^*)$.
If $\mathcal{A}$ is non-exploring on any i.i.d.\ sample $\cD \sim \dP_{\pi_t}(X,Z,Y)$ with probability at least $1 - \delta_t$ for all $t \in \bN$, then
\begin{equation}
\Pr[\pi_T \neq \pi^{*}] > 1 - \sum_{t=0}^T \delta_t \quad \text{for any } T \in \bN \eqp
\end{equation}
\end{proposition}
\begin{proof}
  At each step we have
  \begin{equation*}
  \Pr[\pi_t = \pi^*]
    = \Pr[W_0(\pi_t) = W_0(\pi^*)]
   \le \Pr[W_0(\pi_t) \supset W_0(\pi^*)]
   \le \delta_t + \Pr[\pi_{t-1} = \pi^*] \eqp
 \end{equation*}
 By the assumption that $\pi_0 \ne \pi^*$, we recursively get $\prob[\pi_t = \pi^*] \le \sum_{i=0}^t \delta_i$ which concludes the proof.
\end{proof}
We can thus conclude that, for error based learning algorithms under no fairness constraints, learning within deterministic threshold policies is guaranteed to fail.
Even though the optimal policy lies within the set of deterministic threshold policies, it cannot easily be approximated within this set starting from a suboptimal predictive model.
\begin{figure}
\centering
\begin{tikzpicture}[scale=0.5,decoration={triangles, shape size=0.5mm}]
\draw[thick, MidnightBlue, fill=MidnightBlue!20] plot[smooth cycle, tension=.6] coordinates {(-3.5,0.5) (-3,2.5) (-1,3.5) (1.5,3) (4,3.5) (5,2.5) (5,0.5) (2.5,-2) (0,-0.5) (-3,-2)};
\draw[thick, Orange] plot[smooth, tension=.6] coordinates {(-3, -1) (-1, 1) (1.5, 0.5) (4, 1.6)};
\draw[thick, red, postaction={draw,decorate}] plot[smooth, tension=.6] coordinates {(-1, 1.13) (-0.3, 1.08) (1.5, 0.6) (3.5, 1.4)};
\draw[thick, ForestGreen, postaction={draw,decorate}] (-1, 1.13) to[bend left] (3.5, 1.3);
\node[label={\color{ForestGreen}$\pi^*$}] at (3.5, 1.3) {{\color{ForestGreen}$\bigstar$}};
\node at (2, 4) {\color{MidnightBlue}stochastic policies};
\node at (2, -2.5) {\color{Orange}deterministic policies};
\draw[thick, Orange, ->, >=latex] (1, -2) to[bend left] (1.2, 0.5);
\node[label={[yshift=-0.8cm]$\pi_0$}] at (-1, 1) {{\large $\times$}};
\end{tikzpicture}
\caption{This figure illustrates how it can be impossible to find the optimal policy when the allowed set of policies is restricted to deterministic decision rules.}
\label{fig:impossibility}
\end{figure}
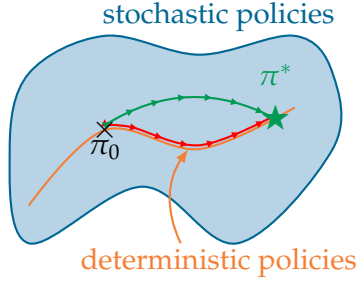

Figure~\ref{fig:impossibility} illustrates that, even though the optimal policy $\pi^*$ is deterministic, when starting from a deterministic initial policy $\pi_0$ (black cross), we cannot iteratively reach $\pi^*$ (green star) when updating solely within deterministic policies (red line following the orange line of deterministic policies).
It is necessary to deploy stochastic policies (blue area) along the way to then be able to converge to the optimal policy (along the green line).
We will introduce such ``exploring policies'' after our final impossibility result for error based learning algorithms.

\begin{corollary}\label{cor:errorbased}
A deterministic threshold policy $\pi \ne \pi^*$ with $\Pr[\pi(x, z) \ne y] = 0$ under $\dP$ will fail to converge to $\pi^*$ under an error based learning algorithm for the underlying predictive model with probability $1$.
\end{corollary}
\begin{proof}
Since error based learning algorithms lead to non-exploring policies whenever
\begin{equation*}
\sum_{(x,z,y) \in \cD} \B{1}[\pi(x,z) \ne y] = 0 \eqc
\end{equation*}
using the assumption $\Pr[\pi(x,z) \ne y] = 0$, we can use Proposition~\ref{prop:limits}
with $\delta_t = 0$ for all $t \in \bN$.
\end{proof}
While we have focused on deterministic threshold rules, our results readily generalize to \emph{all} deterministic policies.
An arbitrary deterministic policy $\pi$ can always be written as a threshold rule $\pi_Q$ as in eq.~\eqref{eq:detthreshQ} with $Q(Y = 1 \given x, z) = \B{1}[\pi(D = 1 \given x, z) = 1]$.
To conclude, if we can only observe the outcomes of previous decisions taken by a deterministic initial policy $\pi_0$, these outcomes may be insufficient to find the (fair) deterministic decision rule that maximizes utility.

\subsection{Stochastic policies}
\label{sec:exploring}

A naive but instructive way to overcome the undesirable behavior exhibited by deterministic policies discussed in the previous section, is to fully randomize initial decisions, i.e., ${\pi_0}(D=1 \given x,z) = \nicefrac{1}{2}$ for all $x,z$.
It readily follows from eq.~\eqref{eq:imperfect-p} that then $\dP_{\pi_0} = \dP$.
Hence, if the hypothesis class of predictive models $\Qcal$ is rich enough, we could learn the optimal policy $\pi^{*}$ from data gathered under $\pi_0$.
In practice, fully randomized initial policies are unacceptable in terms of utility or unethical---it would entail releasing defendants by a coin flip.
Fortunately, we will show next that full randomization is not required to learn the optimal policy.
We only need to choose an initial policy $\pi_0$ such that $\pi_0(D = 1 \given x, z) > 0$ on any measurable subset of $\xspace \times \zspace$ with positive probability under $\dP$, a requirement that is more acceptable for the decision maker in terms of initial utility.
We refer to any policy with this property as an \emph{exploring} policy.
A policy $\pi$ is exploring, if and only if the true distribution $\dP$ is absolutely continuous with respect to the induced distribution $\dP_{\pi}$.
This means the data collection distribution must not ignore regions where the true distribution puts mass.
We note that this condition does not strictly require randomness, but could be achieved by a pre-determined process, e.g., ``$d=1$ for every $n$-th decision''.
For an exploring policy $\pi_0$, we can compute the utility in eq.~\eqref{eq:utility} and the group benefits for $z\in \{0,1\}$ via inverse propensity score weighting
\begin{align}
\begin{split}\label{eq:UtilityStoch}
  u_{\dP_{\pi_0}}(\pi, \pi_0)
  &:= \E_{\substack{x, z, y \sim \dP_{\pi_0} \\ d \sim \pi(x, z)}}
  \Bigl[\frac{d (y-c)}{\pi_0(D = 1 \given x, z)}\Bigr] \eqc \\
  b_{\dP_{\pi_0}}^z(\pi, \pi_0)
  &:= \E_{\substack{x, z, y \sim \dP_{\pi_0} \\ d \sim \pi(x, z)}}
  \Bigl[ \frac{f(d, y)}{\pi_0(D = 1 \given x, z)} \Bigr] \eqp
\end{split}
\end{align}
Crucially, even though $u_{\dP}(\pi) = u_{\dP_{\pi_0}}(\pi, \pi_0)$ and $b_{\dP}^z(\pi) = b_{\dP_{\pi_0}}^z(\pi, \pi_0)$, the expectations are with respect to the induced distribution $\dP_{\pi_0}(X, Z, Y)$, yielding the following positive result.
\begin{proposition}\label{prop:positive}
Let $\Pi$ be the set of exploring policies and let $\pi_0 \in \Pi \setminus \{\pi^*\}$.
Then, the optimal objective value is
\begin{equation*}
   v(\pi^*) = \sup_{\pi \in \Pi \setminus \{\pi^*\}} \Bigl\{u_{\dP_{\pi_0}}(\pi, \pi_0)
  - \frac{\lambda}{2} (b_{\dP_{\pi_0}}^0 (\pi, \pi_0) - b_{\dP_{\pi_0}}^1(\pi, \pi_0))^2 \Bigr\} \eqp
\end{equation*}
\end{proposition}
\begin{proof}
We already know that the supremum is upper bounded by $v(\pi^*)$, i.e., it suffices to construct a sequence of policies $\{\pi_n\}_{n\in \bN_{>0}} \subset \Pi \setminus \{\pi^*\}$ such that $v(\pi_n) \to v(\pi^*)$ for $n \to \infty$.
Using notation from the proof of Proposition~\ref{prop:limitation}, we define
\begin{equation*}
\pi_n(D = 1 \given x,z) :=
\begin{cases}
1 &\text{if } (x, z) \in W_1(\pi^*)\eqc\\
\frac{1}{n} &\text{otherwise}\eqp
\end{cases}
\end{equation*}
It is clear that $\pi_n$ is exploring, i.e., $\pi_n \in \Pi$, for all $n \in \bN_{>0}$ as well as that $\pi_n \ne \pi^*$.
To compute
\begin{equation*}
\lim_{n \to \infty} v_{\dP_{\pi_0}}(\pi_n, \pi_0)
= \lim_{n \to \infty} \Bigl( u_{\dP_{\pi_0}}(\pi_n, \pi_0)
- \frac{\lambda}{2} \bigl(b_{\dP_{\pi_0}}^0(\pi_n, \pi_0) - b_{\dP_{\pi_0}}^1(\pi_n, \pi_0)\bigr)^2 \Bigr)
\end{equation*}
we look at the individual limits.
For the utility we have
\begin{align*}
\lim_{n \to \infty} u_{\dP_{\pi_0}}(\pi_n, \pi_0)
&=
\lim_{n \to \infty}
    \E_{x, z, y \sim \dP_{\pi_0}(X, Z, Y)}
      \left[
        \frac{\pi_n(D = 1 \given x, z)}{\pi_0(D = 1 \given x, z)} (y - c)
      \right] \\
&=
\int_{W_1(\pi^*)} \frac{\dP(Y=1 \given x, z) - c}{\pi_0(D=1 \given x, z)}\, d\dP_{\pi_0}(x,z)\; +
\\
&\quad\;
\lim_{n\to \infty} \frac{1}{n}
\usub{=: C_1 \text{ with } |C_1| < \infty \text{ for any given exploring } \pi_0 \in \Pi}{\int_{W_1(\pi^*)^{\complement}} \frac{\dP(Y=1 \given x, z) - c}{\pi_0(D=1 \given x, z)}\, d\dP_{\pi_0}(x,z)}\\
&= \int_{W_1(\pi^*)} (y-c)\, d\dP(x,z,y) + \lim_{n\to \infty} \frac{C_1}{n}\\
&= u_{\dP}(\pi^*) \eqp
\end{align*}
Similarly, for the benefit terms that are linear in both arguments, such as $f(d, y) = d$, we have for $z \in \{0,1\}$
\begin{align*}
\lim_{n \to \infty} b_{\dP_{\pi_0}}^z(\pi_n, \pi_0)
&= \E_{x, y \sim \dP_{\pi_0}(X, Y \given z)}
          \left[
            \frac{f(\pi_n(D = 1 \given x, z), y)}{\pi_0(D = 1 \given x, z)}
          \right]
\\
&= \int_{W_1(\pi^*)} \frac{f(1, \dP(Y=1 \given x, z))}{\pi_0(D=1 \given x, z)}\, d\dP_{\pi_0}(x \given z)\; +
\\
&\quad\;
\lim_{n\to \infty} \frac{1}{n}
\usub{=: C_2^z \text{ with } |C_2^z| < \infty \text{ for any given exploring } \pi_0 \in \Pi}{\int_{W_1(\pi^*)^{\complement}} \frac{f(1, \dP(Y=1 \given x, z))}{\pi_0(D=1 \given x, z)}\, d\dP_{\pi_0}(x \given z)}\\
&= \int_{W_1(\pi^*)} f(1, y)\, d\dP(x, y \given z) + \lim_{n\to \infty} \frac{C_2^z}{n}\\
&= b_{\dP}^z(\pi^*) \eqp
\end{align*}
Because all the limits are finite, via the rules for sums and products of limits we get
\begin{align*}
  \lim_{n \to \infty} v_{\dP_{\pi_0}}(\pi_n, \pi_0)
 &= \lim_{n \to \infty} u_{\dP_{\pi_0}}(\pi_n, \pi_0) - \frac{\lambda}{2} (\lim_{n \to \infty} b_{\dP_{\pi_0}}^0(\pi_n, \pi_0) - \lim_{n \to \infty} b_{\dP_{\pi_0}}^1(\pi_n, \pi_0))^2 \\
 &= u_{\dP}(\pi^*) - \frac{\lambda}{2} (b_{\dP}^0(\pi^*) - b_{\dP}^1(\pi^*))^2 \\
 &= v_{\dP}(\pi^*) \eqp
\end{align*}
\end{proof}

This shows that---unlike within deterministic threshold models---within exploring policies we can learn the optimal policy using only data from an induced distribution.
Finally, we would like to highlight that not all exploring policies may be (equally) acceptable to society.
For example, in lending scenarios without fairness constraints (i.e., $\lambda=0)$, it may appear wasteful to deny a loan with probability greater than zero to individuals who are believed to repay by the current model.
In those cases, one may like to consider exploring policies that, given sufficient evidence, decide $d=1$ deterministically, i.e., $\pi_0(D = 1 \given x, z) = 1$ for some values of $x, z$.
We will operationalize this notion in \secref{sec:algorithm} as what we call the \emph{semi-logistic policy}.
Other settings, like the criminal justice system, call for a more general discussion about the ethics of non-deterministic decision making.

\section{How to learn exploring policies}
\label{sec:algorithm}

\begin{algorithm}[t!]
\caption{\textsc{ConsequentialLearning}: train a sequence of policies $\pi_{\btheta_t}$ of increasing $v_{\dP}(\pi_{\btheta_t})$.}
\label{algo:SGDpolicylearning}
  \begin{algorithmic}[1]
    \Input cost $c$, time steps $T$, decisions $N$, iterations $M$, minibatch size $B$, penalty $\lambda$, learning rate $\alpha$
        \State $\btheta_0 \gets \Call{InitializePolicy}{{}}$
        \For{$t = 0, \ldots, T-1$} \Comment{time steps}
            \State $\cD^t \gets \Call{CollectData}{{\btheta_t, N}}$
            \State $\btheta_{t+1} \gets \Call{UpdatePolicy}{{\btheta_t, \cD^t, M, B, \alpha}}$
        \EndFor
        \State \Return $\{\pi_{\btheta_{t}}\}_{t=0}^{T}$
    \Statex
    \Function{CollectData}{$\btheta$, $N$}
        \State $\cD \gets \emptyset$
        \For{$i = 1, \ldots, N$} \Comment{$N$ decisions}
            \State $(x_i, z_i) \sim \dP(X,Z)$ and $d_i \sim \pi_{\btheta}(x_i,z_i)$
            \If{$d_i=1$} \Comment{positive decision}
                \State $\cD \gets \cD \cup \{(x_i,z_i,y_i)\}$ with $y_i \sim \dP(Y\given x_i, z_i)$
            \EndIf
        \EndFor
        \State \Return $\cD$ \Comment{data observed under $\pi_{\btheta}$}
    \EndFunction
    \Statex
    \Function{UpdatePolicy}{$\btheta'$, $\cD$, $M$, $B$, $\alpha$}
        \State $\btheta^{(0)} \gets \btheta'$
        \For{$j=1,\ldots, M$} \Comment{iterations}
        \State $\cD^{(j)} \gets \Call{Minibatch}{{\cD}, B}$ \Comment{sample minibatch}
        \State $\nabla \gets 0$, $n_j \gets 0$
        \For{$(x, z, y) \in \cD^{(j)}$} \Comment{accumulate gradients}
                \State $d \sim \pi_{\btheta^{(j)}}(x, z)$
                \If{$d = 1$}
                  \State $n_j \gets n_j + 1$
                  \State $
                    \nabla \gets \nabla +
                    \nabla_{\btheta} v(\pi_{\btheta}, \pi_{\btheta'}) |_{\btheta=\btheta^{(j)}}$
                \EndIf
         \EndFor
         \State $\btheta^{(j+1)} \gets \btheta^{(j)} + \alpha \, \frac{\nabla}{n_j}$
        \EndFor
        \State \Return $\btheta^{M}$
    \EndFunction
  \end{algorithmic}
\end{algorithm}

In this section, we exemplify Proposition~\ref{prop:positive} via a simple, yet practical, gradient-based algorithm to find the solution to eq.~\eqref{eq:policy_learning} within a (differentiable) parameterized class of exploring policies $\Pi(\Theta)$ using data gathered by a given, already deployed, exploring policy $\pi_0$.
To this end, we consider a class of parameterized exploring policies $\Pi(\Theta)$ and we aim to find the policy $\pi_{\btheta^*} \in \Pi(\Theta)$ that solves the optimization problem in eq.~\eqref{eq:policy_learning}.

We use stochastic gradient ascent (SGA) \citep{kiefer1952stochastic} to learn the parameters of the new policy, i.e.,
\begin{equation*}
  \btheta_{i + 1} = \btheta_{i} + \alpha_{i} \nabla_{\btheta} v_{\dP}(\pi_{\btheta}) |_{\btheta = \btheta_{i}}\eqc
\end{equation*}
where
\begin{equation*}
\nabla_{\btheta} v_{\dP}(\pi_{\btheta}) = \nabla_{\btheta} u_{\dP}(\pi_{\btheta}) - \lambda (b_{\dP}^0(\pi_{\btheta}) - b_{\dP}^1(\pi_{\btheta})) (\nabla_{\btheta} b_{\dP}^0(\pi_{\btheta}) - \nabla_{\btheta}b_{\dP}^1(\pi_{\btheta}))\eqc
\end{equation*}
and $\alpha_i > 0$ is the learning rate at step $i \in \bN$.
With the reweighting from eq.~\eqref{eq:UtilityStoch} and the log-derivative trick \citep{williams1992simple}, we can compute the gradient of the utility and the benefits as
\begin{align}
\begin{split}\label{eq:gradient-pi0}
  \nabla_{\btheta} u_{\dP}(\pi_{\btheta})
  &= \E_{\substack{x, z, y \sim \dP_{\pi_0} \\ d \sim \pi_{\btheta}(x, z)}}
  \Big[ \frac{d\, (y - c) \nabla_{\btheta}\log \pi_{\btheta} }{\pi_0(D = 1 \given x, z)} \Big] \eqc \\
  \nabla_{\btheta} b_{\dP}^z(\pi_{\btheta})
  &= \E_{\substack{x, z, y \sim \dP_{\pi_0} \\ d \sim \pi_{\btheta}(x, z)}}
  \Big[ \frac{f(d,y) \nabla_{\btheta}\log \pi_{\btheta}}{\pi_0(D = 1 \given x, z)} \Big] \eqc
\end{split}
\end{align}
where $\nabla_{\btheta} \log \pi_{\btheta} := \nabla_{\btheta} \log \pi_{\btheta} (D \given x, z)$ is the score function \citep{Hyvarinen05:ScoreMatching}.
Thus, our implementation resembles a REINFORCE algorithm with horizon one.

Note that we can obtain an expression for $\nabla_{\btheta_t} v_{\dP}(\pi_{\btheta_t})$ by simply replacing $\pi_0$ with $\pi_{\btheta_{t-1}}$ in eq.~\eqref{eq:gradient-pi0}.
Thus we can estimate the gradient with samples $(x_i, z_i, y_i)$ from the distribution $\dP_{\pi_{t-1}}$ induced by the previous policy $\pi_{t-1}$, and sample the decisions from the policy under consideration $d_i \sim \pi_{\btheta_t}$.
This yields an unbiased finite sample Monte-Carlo estimator for the gradients
\begin{align}
\begin{split}\label{eq:gradutilestim}
  & \nabla_{\btheta_t} u(\pi_{\btheta_t}, \pi_{\btheta_{t-1}}) \approx \frac{1}{n_{t-1}}
  \sum_{i = 1}^{n_{t-1}} \frac{d_i (y_i - c)}{\pi_{\btheta_{t-1}}(D = 1 \given x_i, z_i)}\, \nabla_{\btheta_t} \log \pi_{\btheta_t}(D = d_i \given x_i, z_i) \eqc
  \\
  & \nabla_{\btheta_t} b^z(\pi_{\btheta_t}, \pi_{\btheta_{t-1}}) \approx \frac{1}{n_{t-1}}
  \sum_{i = 1}^{n_{t-1}} \frac{f(d_i, y_i)}{\pi_{\btheta_{t-1}}(D = 1 \given x_i, z_i)}\, \nabla_{\btheta_t} \log \pi_{\btheta_t}(D = d_i \given x_i, z_i) \eqc
\end{split}
\end{align}
where $n_{t-1}$ is the number of positive decisions taken by $\pi_{\btheta_{t-1}}$.
Here, it is important to notice that, while the decisions by $\pi_{\btheta_{t-1}}$ were actually taken and, as a result, (feature and label) data was gathered under $\pi_{\btheta_{t-1}}$, the decisions $d_i \sim \pi_{\btheta_{t}}$ are just sampled to implement SGA.
The overall policy learning process is summarized in Algorithm~\ref{algo:SGDpolicylearning}, where \textsc{Minibatch}$(\cD, B)$ samples a minibatch of size $B$ from the dataset $\cD$ and \textsc{InitializePolicy}$()$ initializes the policy parameters.

Unfortunately, the above procedure has two main drawbacks.
First, it may require an abundance of data from $\dP_{\pi_0}$, which can be unacceptable in terms of utility if $\pi_0$ is far from optimal.
Second, if $\pi_0(D = 1 \given x, z)$ is small in a region where $\pi_{\btheta}$ often takes positive decisions, one may expect that an empirical estimate of the above gradient will have high variance, due to similar arguments as in weighted inverse propensity scoring \citep{Sutton98:RL}.
On the other hand, in most practical applications updating the model after every single decision is impractical.
Typically, a fixed model will be deployed for a certain period, before it is updated using the data collected within this period.
This is also a natural mode of operation for predictive models in real-world applications.

\begin{figure}
\centering
\begin{minipage}{0.5\textwidth}
\centering
\begin{tikzpicture}%
    \node[anchor=south west,inner sep=0] (image) at (0,0) {%
      \includegraphics[width=\columnwidth]{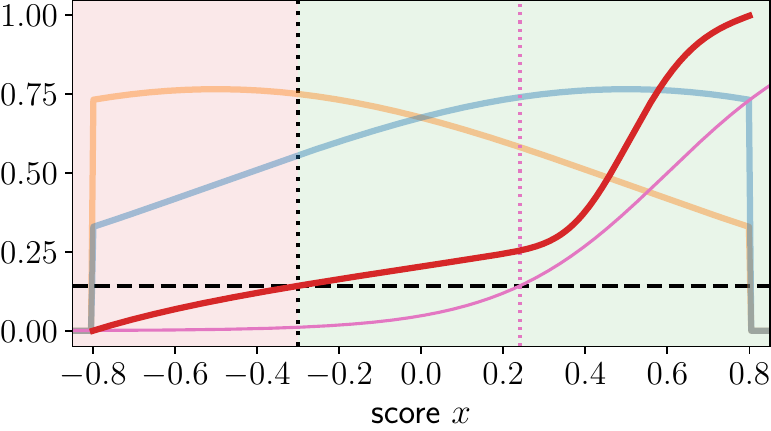}};
    \begin{scope}[x={(image.south east)}, y={(image.north west)}]
        \node at (1.25,0.6) {{\small \textbf{First Setting}}};
    \end{scope}
\end{tikzpicture}\\
\begin{tikzpicture}%
    \node[anchor=south west,inner sep=0] (image) at (0,0) {%
      \includegraphics[width=\columnwidth]{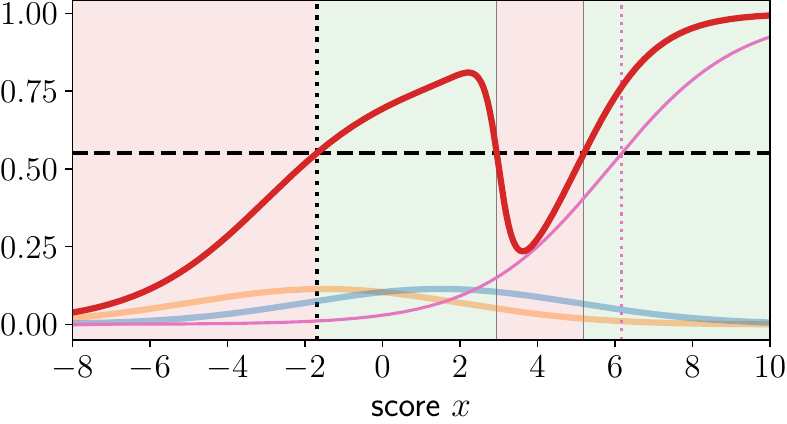}};
    \begin{scope}[x={(image.south east)}, y={(image.north west)}]
        \node at (1.25,0.4) {{\small \textbf{Second Setting}}};
    \end{scope}
\end{tikzpicture}%
\end{minipage}
\hspace{0.5cm}
\begin{minipage}{0.3\columnwidth}
\includegraphics[width=\columnwidth]{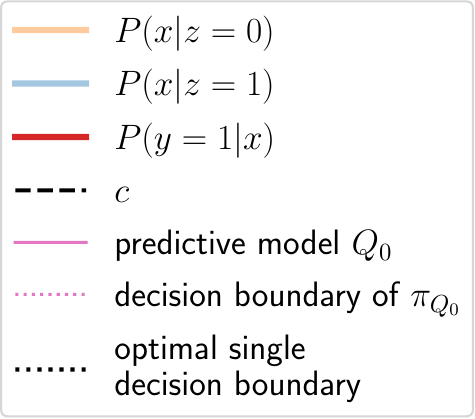}%
\end{minipage}
\caption{Two synthetic settings.
In red, we show $\dP(Y = 1\given x)$, where the score $x$ is drawn from different distributions for the two groups (blue/orange).
For given $c$ (black, dashed), the optimal policy decides $d=1$ ($d=0$) in the shaded green (red) regions.
The vertical black, dotted line shows the best policy achievable with a single threshold on $x$.
In pink, we show a possible imperfect logistic predictive model and its corresponding (suboptimal) threshold in $x$.}
\label{fig:synthetic-setting}
\end{figure}

To overcome these drawbacks, we build two types of sequences of policies $\{ \pi_{\btheta_t} \}_{t=0}^{T}$:
\begin{enumerate}[label=\alph*)]
  \item the \emph{iterative sequence} $\pi_{t+1} := \mathcal{A}(\pi_t, \cD^t)$ with $\cD^t \sim \dP_{\pi_t}(X,Z,Y)$, where only the data gathered by the immediately previous policy are used to update the current policy;
  \item the \emph{aggregated sequence} $\pi_{t+1} := \mathcal{A}(\pi_t, \bigcup_{i=0}^t \cD^i)$ with $\cD^i \sim \dP_{\pi_i}(X,Z,Y)$, where the data gathered by all previous policies are used to update the current policy.
\end{enumerate}

\xhdr{Remarks}
Note that in Algorithm~\ref{algo:SGDpolicylearning} we learn each policy $\pi_t$ only using data from the previous policy $\pi_{t-1}$.
This may readily be generalized to a mix of multiple previous policies $\pi_{t'}$ in eq.~\eqref{eq:gradutilestim}.
Averaging multiple gradient estimators for several $t' < t$ is again an unbiased gradient estimator.
To reduce variance, in practice one may consider recent policies $\pi_{t'}$ most similar to $\pi_t$.

The way in which we use weighted sampling to estimate the above gradients closely relates to the concept of weighted inverse propensity scoring (wIPS), commonly used in counterfactual learning \citep{Bottou13:Counterfactual,Swaminathan15:CRM}, off-policy reinforcement learning \citep{Sutton98:RL}, and contextual bandits \citep{Langford08:ES}.
However, a key difference is that, in wIPS, the labels $y$ are always observed.
As an example, in the case of counterfactual learning one may interpret $\pi_0(x,z)$ in eq.~\eqref{eq:imperfect-p} as a treatment assignment mechanism in a randomized controlled trial.
Under this interpretation, the two most prominent differences with respect to the literature become
apparent. First, we do not observe outcomes in the control group. Second, in observational studies for treatment effect estimation \citep{Rubin05:PO}, one usually estimates
the direct causal effect of $d$ on $y$, i.e., $\dP(Y \given do(D=d'), x, z)$, in the presence of confounders
$x, z$ that affect both $d$ and $y$. This could be evaluated in a (partially) randomized controlled trial, where wIPS also comes in naturally \citep{Pearl2009}.
In contrast, in our setting, the true label $y$ is independent of the decision $d$ and we estimate the conditional $\dP(Y \given x,z)$ using data from the induced distribution $\dP_{\pi_0}(X,Z) \propto \dP(X,Z) \pi_0(D = 1\given x,z)$.
With exploring policies, we obtain indirect access to the true data distribution $\dP(x,z)$ (positivity), and thus to an unbiased estimator of the conditional distribution $\dP(Y\given x,z)$ (consistency).

Despite this difference, we believe that recent advances to reduce the variance of the gradients in weighted inverse propensity scoring, such as
clipped-wIPS \citep{Bottou13:Counterfactual}, self-normalized estimator \citep{Swaminathan15:SEC}, or doubly robust estimators \citep{Dudik11:DoublyRobust},
may also be applicable to our setting.

Finally, we opt for the simple SGA approach on \mbox{(semi-)logistic} policies over, e.g., contextual bandits algorithms, because it provides a direct and fairer comparison with commonly used prediction based decision policies (e.g., logistic regression), also often trained via SGA.

While our algorithm works for any differentiable class of exploring policies, here we consider two examples of exploring policy classes in particular.

\xhdr{Logistic policy}
Our first concrete parameterization of $\pi_{\btheta}$, a \emph{logistic policy} is given by
\begin{equation*}
    \pi_{\btheta}(D = 1\given x, z) = \sigma(\bphi(x,z)^{\top} \btheta) \in (0,1) \eqc
\end{equation*}
where $\sigma(a) := \frac{1}{1+\exp(-a)}$ is the logistic function, $\btheta \in \Theta \subset \bR^m$ are the model parameters, and
$\bphi: \bR^d \times \{0,1\} \to \bR^m$ is a fixed feature map.
Note that any logistic policy is an exploring policy and we can analytically compute its score function $\nabla_{\btheta_t} \log \pi_{\btheta_t} (D=1\given x, z)$ as
\begin{equation*}
 \nabla_{\btheta_t} \log(\sigma(\bphi_i^{\top} \btheta_t))
 = \frac{\bphi_i}{1 + e^{\bphi_i^{\top} \btheta_t}} \in \bR^m \eqc
\end{equation*}
where $\bphi_i := \bphi(x_i, z_i)$.
Using this expression, we can rewrite the empirical estimator for the gradient in eq.~\eqref{eq:gradutilestim}
\begin{align*}
  &\nabla_{\btheta_t} u(\pi_{\btheta_t}, \pi_{\btheta_{t-1}}) \approx \frac{1}{n_{t-1}}
  \sum_{i = 1}^{n_{t-1}} \frac{1 + e^{-\bphi_i^{\top} \btheta_{t-1}}}{1 + e^{\bphi_i^{\top} \btheta_t}}\, d_i\, (y_i - c)\, \bphi_i \eqc
  \\
  & \nabla_{\btheta_t} b^z(\pi_{\btheta_t}, \pi_{\btheta_{t-1}}) \approx \frac{1}{n_{t-1}}
  \sum_{i = 1}^{n_{t-1}} \frac{1 + e^{-\bphi_i^{\top} \btheta_{t-1}}}{1 + e^{\bphi_i^{\top} \btheta_t}}\, f(d_i, y_i)\, \bphi_i \eqp
\end{align*}
Given the above expression, we have all the necessary ingredients to implement Algorithm~\ref{algo:SGDpolicylearning}.

\xhdr{Semi-logistic policy}
As discussed in the previous section, randomizing decisions may be questionable in certain practical scenarios.
For example, in loan decisions, it may appear wasteful for the bank and contestable for the applicant to deny a loan with probability greater than zero to individuals who are believed to repay by the current model.
In those cases, one may consider the following modification of the logistic policy, which we refer to as \emph{semi-logistic policy}:
\begin{equation*}
    \pitil_{\btheta}(D = 1\given x, z) =
    \begin{cases}
    1 & \text{ if } \bphi(x, z)^{\top} \btheta \ge 0 \eqc \\
    \sigma(\bphi(x, z)^{\top} \btheta) &\text{ if } \bphi(x, z)^{\top} \btheta < 0 \eqp
    \end{cases}
\end{equation*}
Similarly as in the logistic policy, we can compute the score function analytically as:
\begin{equation*}
    \nabla_{\btheta} \log \pitil_{\btheta}(D = 1 \given x, z) =
    \frac{\bphi(x, z)}{1 + e^{\bphi(x, z)^{\top}\btheta}} \, \B{1}[\bphi(x, z)^{\top}\btheta < 0] \eqc
\end{equation*}
and use this expression to compute an unbiased estimator for the gradient in eq.~\eqref{eq:gradutilestim} as:
\begin{align*}
  \nabla_{\btheta_t} u(\pi_{\btheta_t}, \pi_{\btheta_{t-1}})
  &\approx \frac{1}{n_{t-1}}
  \sum_{\substack{i=1 \\ \bphi_i^{\top}\btheta_t < 0}}^{n_{t-1}} \frac{d_i\, (y_i - c)\, \bphi_i} {1 + e^{\bphi_i^{\top} \btheta_t}}
  \times
  \begin{cases}
      1 &\text{ if } \bphi_i^{\top} \btheta_{t-1} \ge 0 \eqc \\
      (1 + e^{-\bphi_i^{\top}\btheta_{t-1}}) &\text{ if } \bphi_i^{\top} \btheta_{t-1} < 0 \eqp
  \end{cases}
  \\
  \nabla_{\btheta_t} b^z(\pi_{\btheta_t}, \pi_{\btheta_{t-1}})
  &\approx \frac{1}{n_{t-1}}
  \sum_{\substack{i=1 \\ \bphi_i^{\top}\btheta_t < 0}}^{n_{t-1}} \frac{f(d_i, y_i)\, \bphi_i} {1 + e^{\bphi_i^{\top} \btheta_t}}
  \times
  \begin{cases}
      1 &\text{ if } \bphi_i^{\top} \btheta_{t-1} \ge 0 \eqc \\
      (1 + e^{-\bphi_i^{\top}\btheta_{t-1}}) &\text{ if } \bphi_i^{\top} \btheta_{t-1} < 0 \eqp
  \end{cases}
\end{align*}
Note that the semi-logistic policy is an exploring policy and thus satisfies the assumptions of Proposition~\ref{prop:positive}. Finally, in all our experiments, we directly work with the available features $x$ as inputs and add a constant offset, i.e., $\phi(x, z) = (1, x)$.

\section{Experiments}
\label{sec:experiments}

We learn a sequence of policies $\{ \pi_{\btheta_t} \}_{t=1}^{T}$ using the following strategies:

\begin{description}
\item[Optimal:] decisions are taken by the optimal deterministic threshold rule $\pi^{*}$ given by eq.~\eqref{eq:detthresh}, i.e., $\pi_t = \pi^{*}$ for all $t$.
It can only be computed when the ground truth conditional $\dP(Y \given x, z)$ is known.
\item[Deterministic:] decisions are taken by deterministic threshold policies $\pi_t = \pi_{Q_t}$, where $Q_t$ are logistic predictive models maximizing label likelihood trained either in an iterative or aggregate sequence.
\item[Logistic:] decisions are taken by logistic policies $\pi_t = \pi_{\btheta_t}$ trained via Algorithm~\ref{algo:SGDpolicylearning} either in an iterative or aggregate sequence.
\item[Semi-logistic:] decisions are taken by semi-logistic policies $\pitil_t = \pitil_{\btheta_t}$ trained via Algorithm~\ref{algo:SGDpolicylearning} either in an iterative or aggregate sequence.
\end{description}

It is crucial that while each of the above methods decides over the same set of proposed $\{(x_i, z_i)\}_{i=1}^N$ at each time step $t$, depending on their decisions, they may collect labels for differing subsets and thus receive different amounts of new training data.
During learning, we record the following metrics:\footnote{For readability we only show medians over 30 runs. Figures with 25 and 75 percentiles are in Appendix~\ref{sec:appendix}.}

\begin{description}
\item[Utility:] the utility $u_{\dP}(\pi_t)$ achieved by the current policy $\pi_t$ estimated empirically on a held-out dataset, the \emph{test set}, sampled i.i.d.\ from the ground truth distribution $\dP(X,Z,Y)$.
This is the utility that the decision maker would obtain if they deployed the current policy $\pi_t$ at large in the population.
\item[Effective utility:] the utility realized during the learning process up to time $t$, i.e.,
\begin{equation*}
\hat{u}(t) = \frac{1}{N\cdot t} \sum_{t' \leq t} \sum_{(x_i, z_i, y_i) \in \cD^{t'}} (y_i - c)\eqc
\end{equation*}
where $\cD^{t'}$ are the data in which the policy $\pi_{t'}(x_i, z_i)$ took positive decisions $d_i = 1$ and $N$ is the number of considered examples at each time step $t$.
This is the utility accumulated by the decision maker while learning better policies.
\item[Fairness:] the difference in group benefits between sensitive groups $\Delta b_{\dP}(\pi) := b_{\dP}^0(\pi) - b_{\dP}^1(\pi)$ for disparate impact: $f(d, y) = d$.
A policy fully satisfies the chosen criterion if and only if $\Delta b_{\dP}(\pi) = 0$.
Again, we estimate fairness empirically on the test set and thus measure the level of fairness $\pi_t$ would achieve in the entire population.
\end{description}

Detailed parameter settings for the experiments are explained in \secref{sec:app:parameters}.

\subsection{Experiments on synthetic data}

\begin{figure}
\centering
\def\figheight{4cm}
\includegraphics[width=\textwidth]{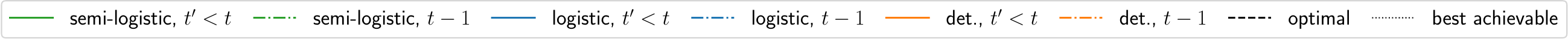}\\
\textbf{First Setting}\\
\includegraphics[height=\figheight]{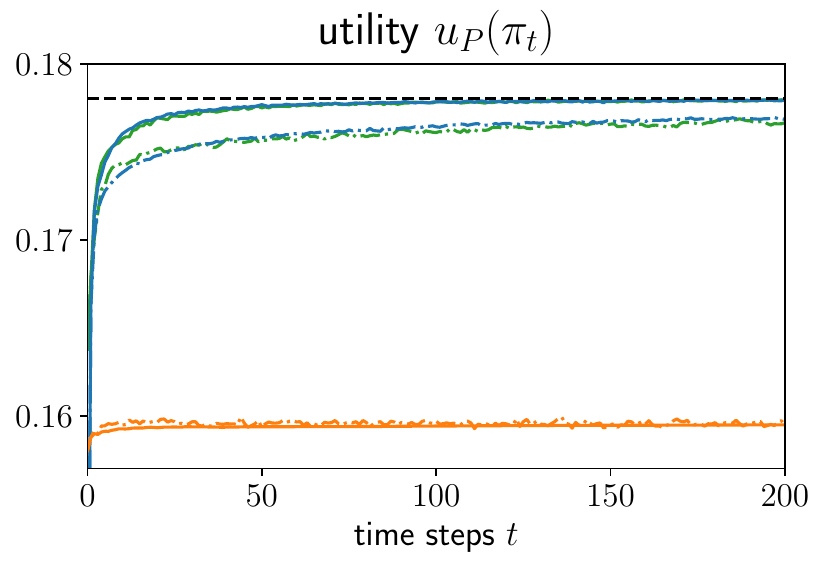}%
\hspace{1cm}
\includegraphics[height=\figheight]{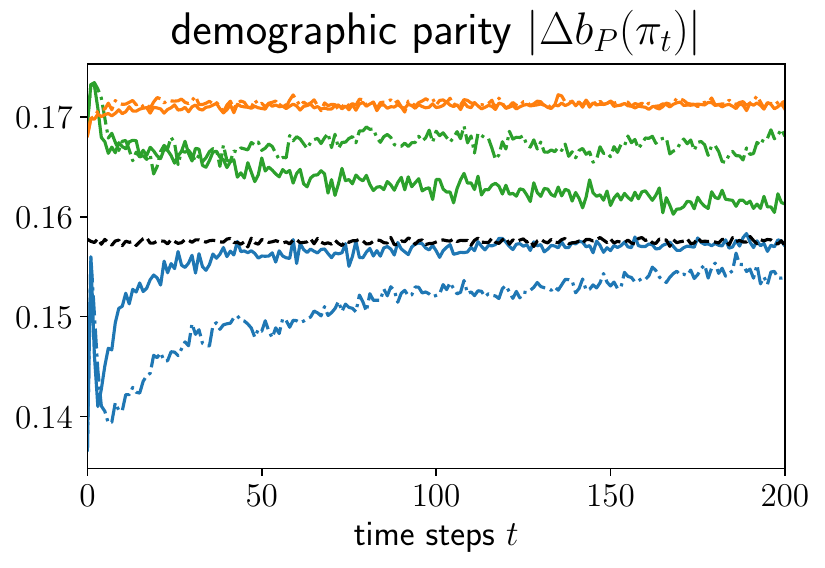}\\%
\textbf{Second Setting}\\
\includegraphics[height=\figheight]{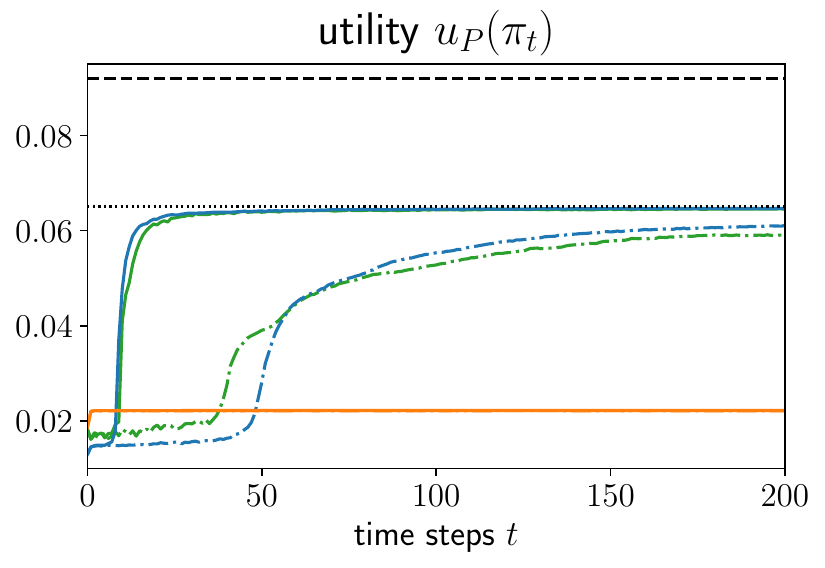}%
\hspace{1cm}
\includegraphics[height=\figheight]{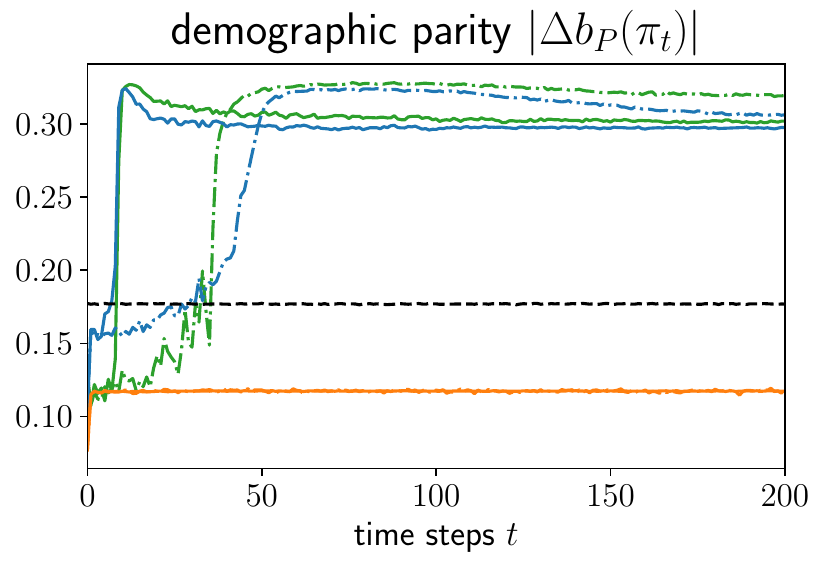}%
\caption{Utility and demographic parity in the synthetic settings of Figure~\ref{fig:synthetic-setting} without enforcing fairness constraints, i.e., $\lambda=0$.}
\label{fig:results-synthetic}
\end{figure}

We assume that there is a single non-sensitive feature $x \in \bR$ per individual---similar to the lending example in \secref{sec:sequential}---and a sensitive attribute $z \in \{0,1\}$.
While $\dP(X \given z=0) \ne \dP(X \given z=1)$, in our experiments the policies only take $x$ as input, and \emph{not} the sensitive attribute, which is only used for the fairness constraints.
We consider two different settings, illustrated in Figure~\ref{fig:synthetic-setting}, where $z \sim \mathrm{Ber}(0.5)$ and the distributions over $x$ differ for the two groups, see Appendix~\ref{sec:appendix}.
In the first setting, the conditional probability $\dP(Y = 1 \given x)$ is strictly monotonic in the score and does not depend on $z$, but is not well calibrated, i.e., not directly proportional to $x$.
In the second setting, the conditional probability $\dP(Y = 1\given x)$ crosses the cost threshold $c$ multiple times resulting in two disjoint intervals for which the optimal decision is $d=1$ (green areas).

Figure~\ref{fig:results-synthetic} summarizes the results for $\lambda = 0$, i.e., without explicitly enforcing fairness constraints.
Our method outperforms prediction based deterministic threshold rules in terms of utility in both settings.
This can be easily understood from the evolution of policies illustrated in Figure~\ref{fig:results_synthetic_evolution} in Appendix~\ref{sec:appendix}.
In the first setting, exploring policies locate the optimal decision boundary, whereas the deterministic threshold rules get stuck, even though $\dP(Y = 1\given x)$ is monotonic in $x$.
In the second setting, our methods explore more and eventually identify the best single threshold at the black vertical dotted line in Figure~\ref{fig:synthetic-setting}.
In contrast, non-exploring deterministic threshold rules converge to a suboptimal threshold at $x\approx 5$, ignoring the left green region.

In the first setting, we also observe that the suboptimal predictive models amplify unfairness beyond the levels exhibited by the optimal policy.
For our approach, levels of unfairness are comparable to or even below those of the optimal policy.
The second setting shows that depending on the ground truth distribution, higher utility can be directly linked to larger fairness violations.
In such cases, our approach allows to explicitly control for fairness.
Results on utility and demographic parity under fairness constraints with different $\lambda$ are shown in Figure~\ref{fig:results-synthetic-final-dp} in Appendix~\ref{sec:appendix}.
In essence, $\lambda$ trades off utility and fairness violations to the point of perfect fairness in the ground truth distribution.

\subsection{Experiments on real data}

\begin{figure}
\centering
\includegraphics[width=\textwidth]{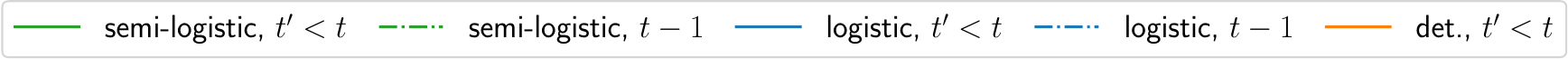}\\
\includegraphics[width=0.32\textwidth]{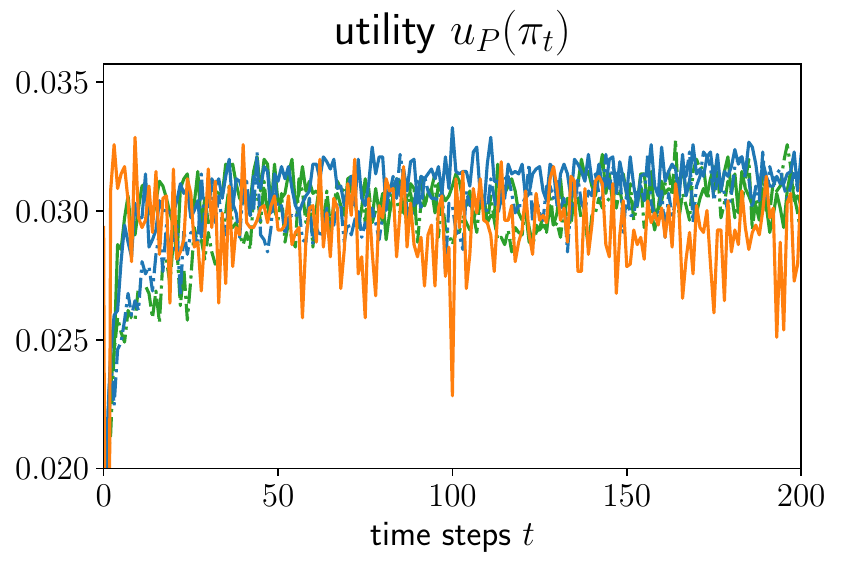}
\includegraphics[width=0.32\textwidth]{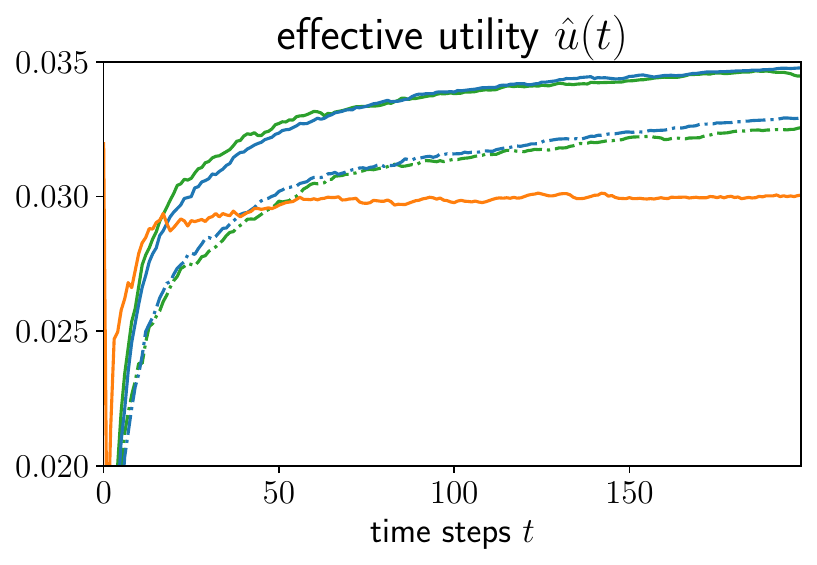}
\includegraphics[width=0.32\textwidth]{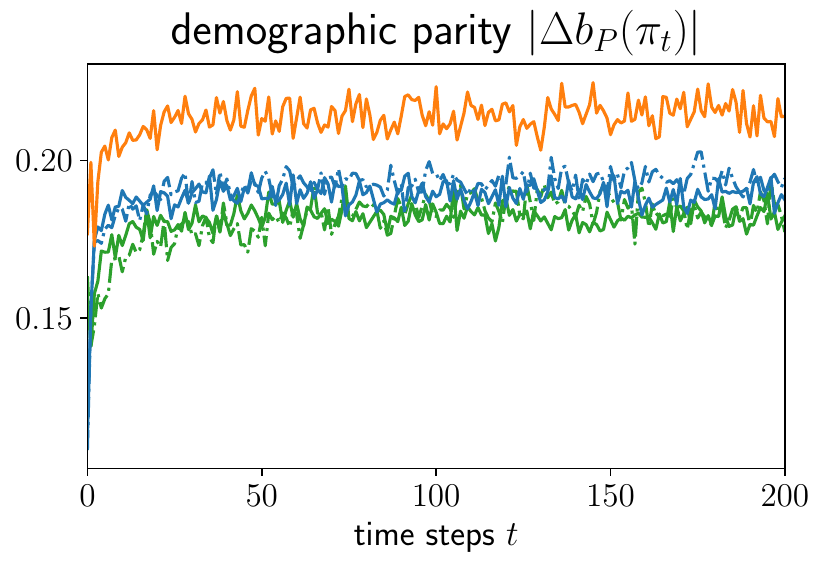}
\caption{Training progress on COMPAS data for $\lambda = 0$, i.e., without fairness constraints.}
\label{fig:results-real-time}
\end{figure}

\begin{figure}
\centering
\def\figheight{4cm}
\def\figspacing{1}
\includegraphics[width=\textwidth]{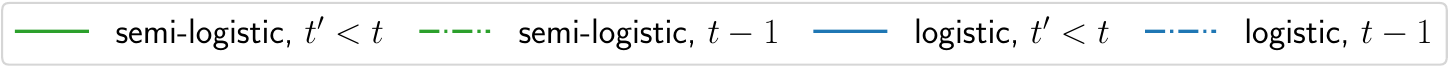}\\
\includegraphics[height=\figheight]{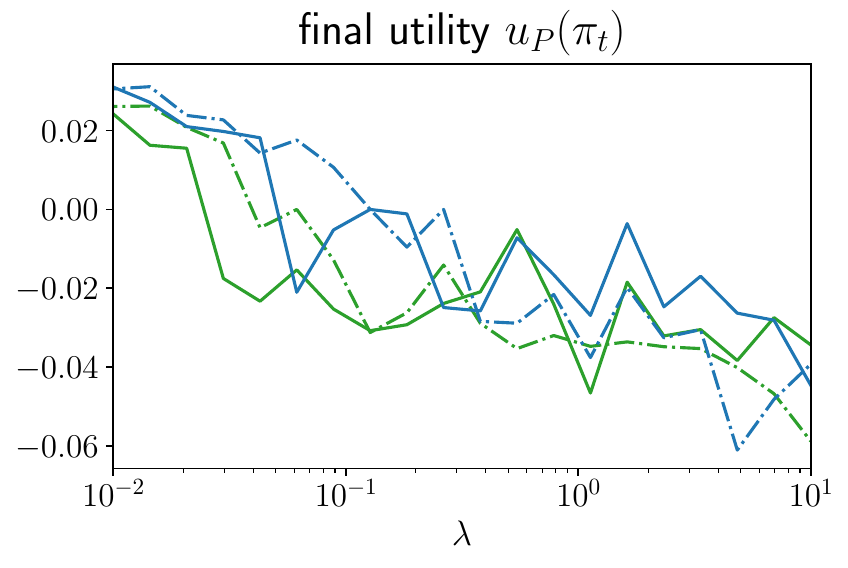}
\hspace{\figspacing cm}
\includegraphics[height=\figheight]{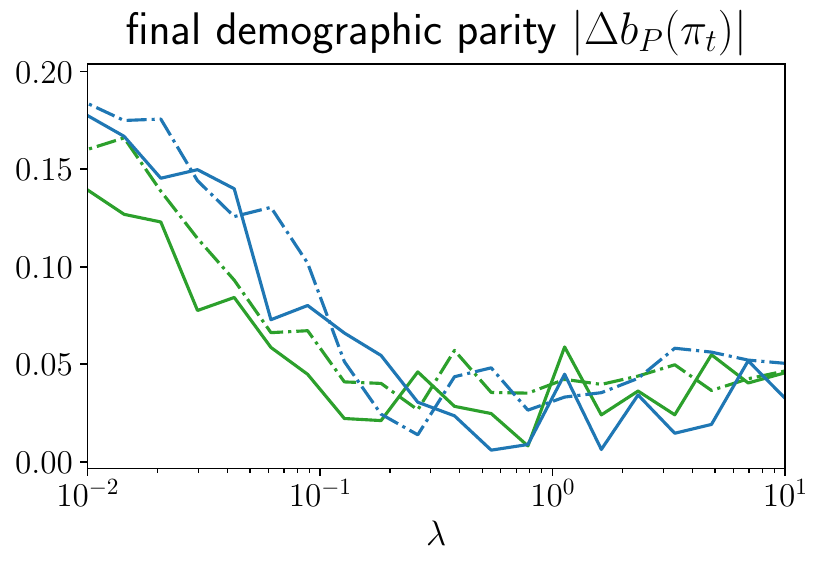}
\caption{Fairness evaluation on COMPAS data for the final ($t=200$) policy as a function of $\lambda$ for demographic parity.
All quantities are estimated on the held-out set.}
\label{fig:results-real-lambda}
\end{figure}

Here, we use the COMPAS recidivism dataset compiled by ProPublica \citep{Angwin2016}, which comprises of information about criminal offenders screened through the COMPAS tool in Broward County, Florida during 2013-2014.
For each offender, the dataset contains a set of demographic features, the criminal history, and the risk score assigned by COMPAS.
Moreover, ProPublica collected whether or not these individuals were rearrested within two years of the screening.
In our experiments, $z \in \{0, 1\}$ indicates whether individuals were identified ``white'', $y$ indicates rearrest, and $d \sim \pi(x, z)$ determines whether an individual is let out on parole.
Again, $z$ is not used as an input.
We use 80\% of the data for training, where at each step $t$, we sample (with replacement) $N$ individuals, and the remaining 20\% as a held-out set to evaluate each learned policy in the population of interest.

We first summarize the results for $\lambda=0$, i.e., without fairness constraints in Figure~\ref{fig:results-real-time}.
A slight initial utility advantage of the deterministic threshold rule is quickly overcome by our exploring policies.
This is best seen when looking at \emph{effective utility}, the average utility accumulated by the decision maker on training data up to time $t$, for which our strategies dominate after $t=100$.
Hence, early exploration not only pays off to eventually be able to take better decisions, but also reaps higher profit during training.
Moreover, all strategies based on exploring policies consistently achieve lower violations of demographic parity than the deterministic threshold rules.
In summary, even without fairness constraints, i.e., in a pure utility maximization setting, exploring policies achieve higher utility and simultaneously reduce unfairness compared to deterministic threshold rules.

In Figure~\ref{fig:results-real-lambda}, we show how utility and demographic parity of the final policy $\pi_{t=200}$ changes as a function of $\lambda$ when constraining demographic parity.
As expected, while we are able to achieve perfect demographic parity, this comes with a drop in utility.
All remaining metrics under fairness constraints are shown in Figure~\ref{fig:results-real-final} in Appendix~\ref{sec:appendix}.
Finally, two remarks are in order.
First, for real-world data we cannot evaluate the optimal policy and do not expect it to reside in our model class.
However, even when logistic models do not perfectly capture the conditional $\dP(Y = 1 \given x)$, our comparisons here are ``fair'' in that all strategies have equal modeling capacity.
Second, we take the COMPAS dataset as our (empirical) ground truth distribution even though it likely also suffered from selective labels.
To learn about the real distribution underlying the dataset, we would need to actually deploy our strategy.

\section{Conclusion}
\label{sec:conclusions}

In this paper, we have analyzed consequential decision making using imperfect predictive models, which are learned from data gathered by potentially biased historical decisions.
First, we have articulated how this approach fails to optimize utility when starting with a non-optimal deterministic policy.
Next, we have presented how directly learning to decide with exploring policies avoids this failure mode while respecting a common fairness constraint.
Finally, we have introduced and evaluated a simple, yet practical gradient-based algorithm to learn fair exploring policies.

Unlike most previous work on fairness in machine learning, which phrases decision making directly as a prediction problem, we argue for a shift from ``learning to predict'' to ``learning to decide''.
In particular, we propose to not simply equate decisions with predictions obtained directly from limited available data, but to remain cognizant of how decisions can affect and interfere with future data collection by continued exploration.
Not only does this lead to improved fairness in this context, but it also establishes connections to other areas such as counterfactual inference, reinforcement learning and contextual bandits.
Within reinforcement learning, it would be interesting to move beyond a static distribution $\dP$ by incorporating feedback from decisions or non-static externalities.
Moreover, since we have shown how shifting focus from learning predictions to learning decisions requires exploration, we hope to stimulate future research on how to explore ethically in different domains.

\bibliographystyle{refstyle}
\bibliography{references}

\clearpage
\appendix

\section{Additional experiments}
\label{sec:appendix}

\begin{figure}
\centering
\def\figheight{3.5cm}
\includegraphics[width=\textwidth]{legend-synthetic-time.pdf}\\
\includegraphics[height=\figheight]{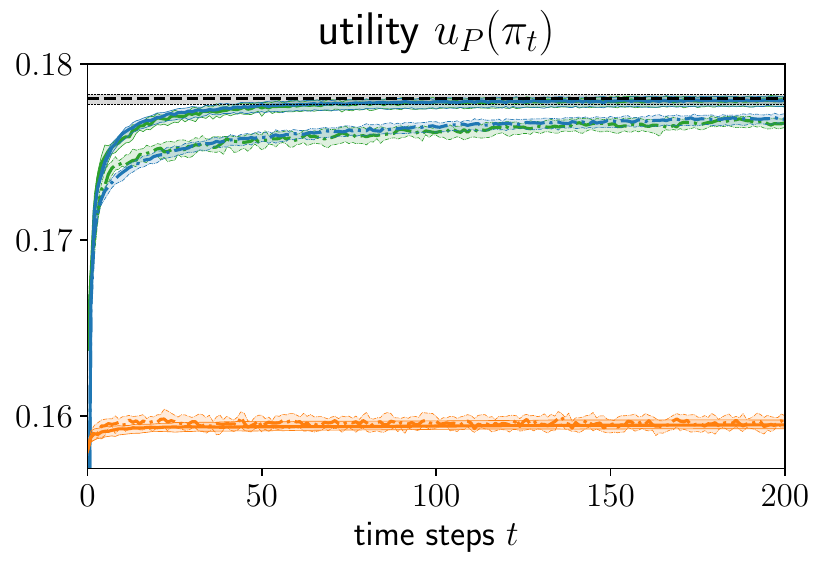}%
\hfill
\includegraphics[height=\figheight]{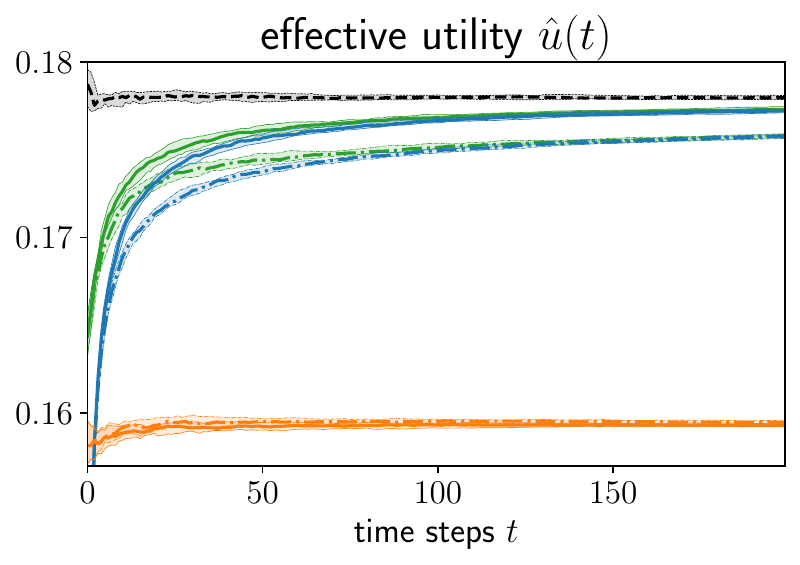}
\hfill
\includegraphics[height=\figheight]{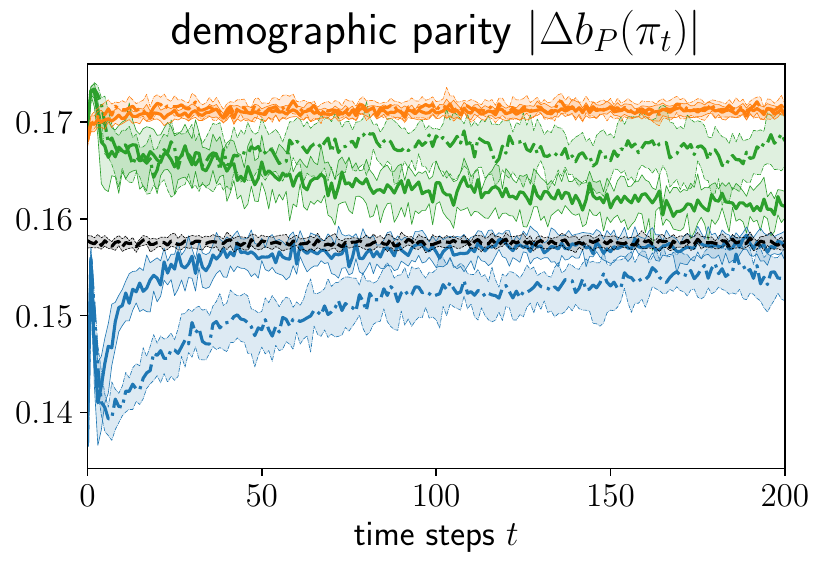}%
\\
\includegraphics[height=\figheight]{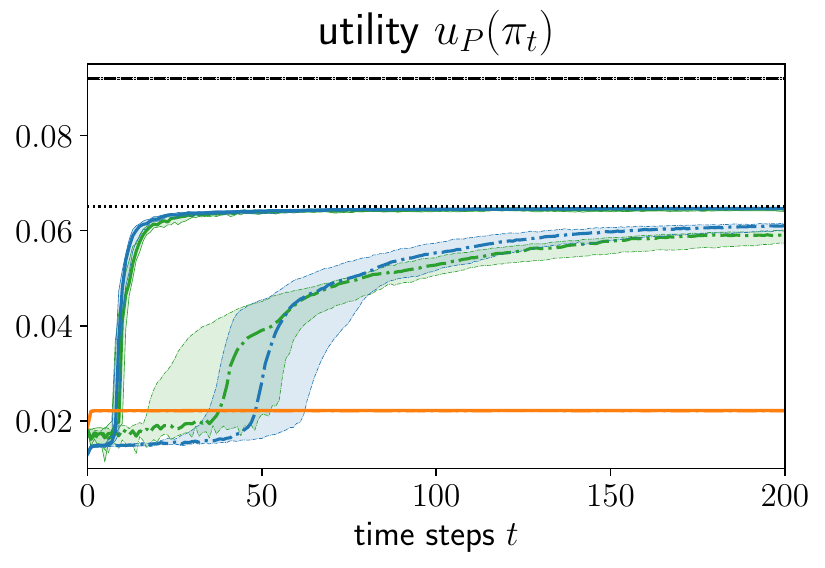}%
\hfill
\includegraphics[height=\figheight]{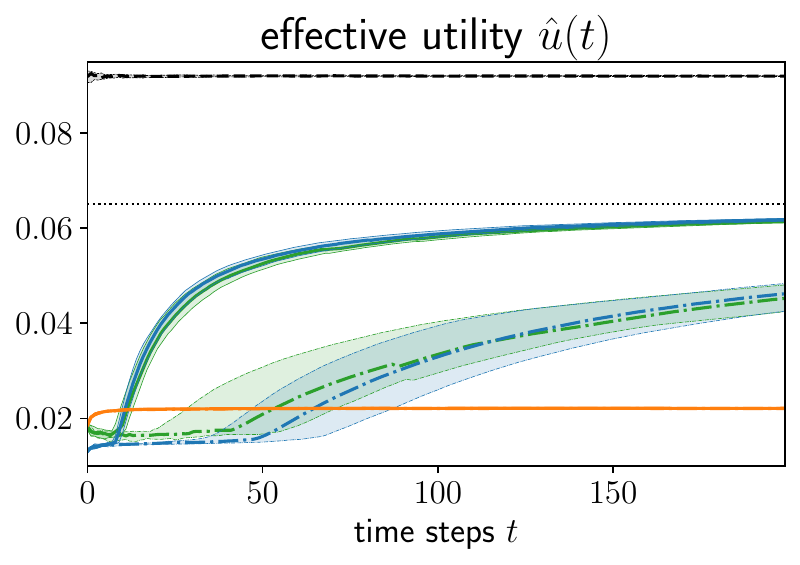}
\hfill
\includegraphics[height=\figheight]{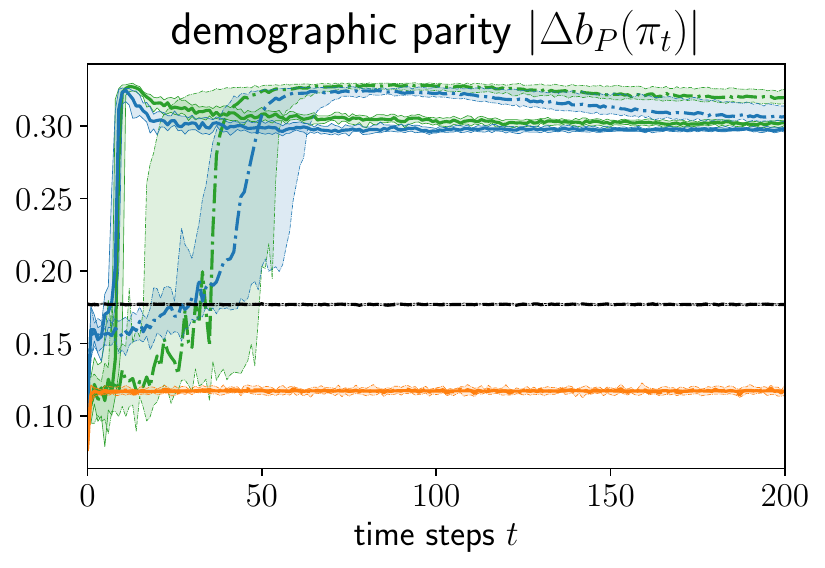}%
\caption{Utility, effective utility, and demographic parity in the synthetic settings of Figure~\ref{fig:synthetic-setting}.}
\label{fig:results-synthetic-errs}
\end{figure}

\begin{figure}
\captionsetup[subfigure]{labelformat=empty}
\captionsetup[subfloat]{farskip=-1mm,captionskip=-0.5mm}
\centering
\def\figwid{0.27}
\subfloat[\tiny\textbf{deterministic, $t'<t$}]{%
\includegraphics[width=\figwid \textwidth]{%
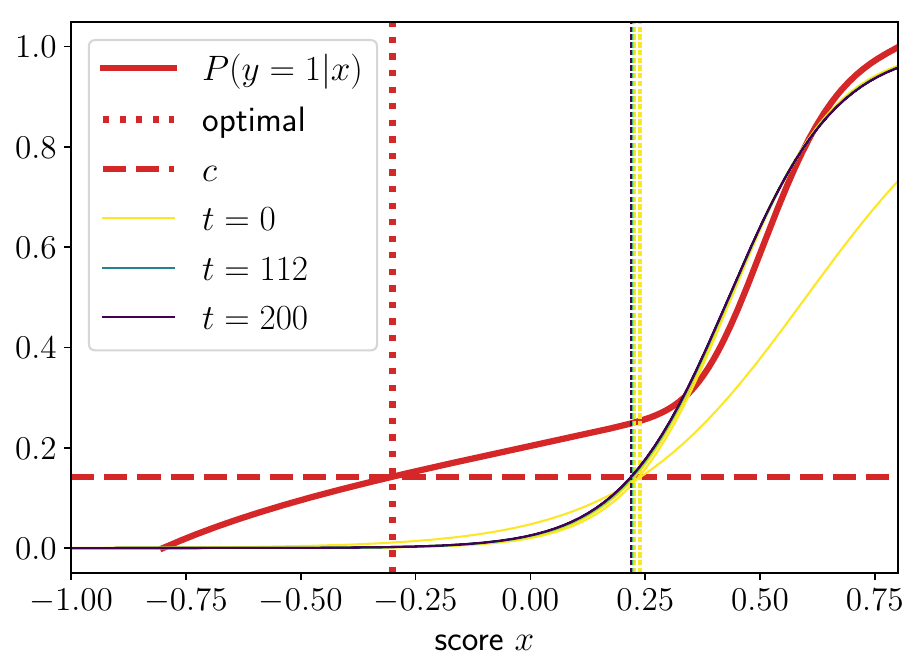}}
\hspace{0.5cm}
\subfloat[\tiny\textbf{deterministic, $t'<t$}]{%
\includegraphics[width=\figwid \textwidth]{%
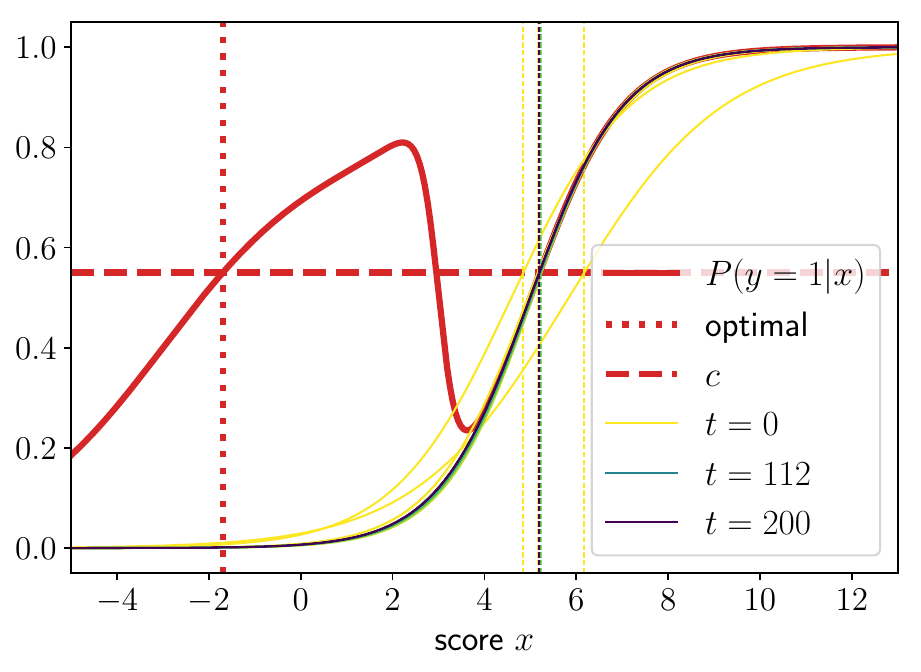}}
\\
\subfloat[\tiny\textbf{deterministic, $t-1$}]{%
\includegraphics[width=\figwid \textwidth]{%
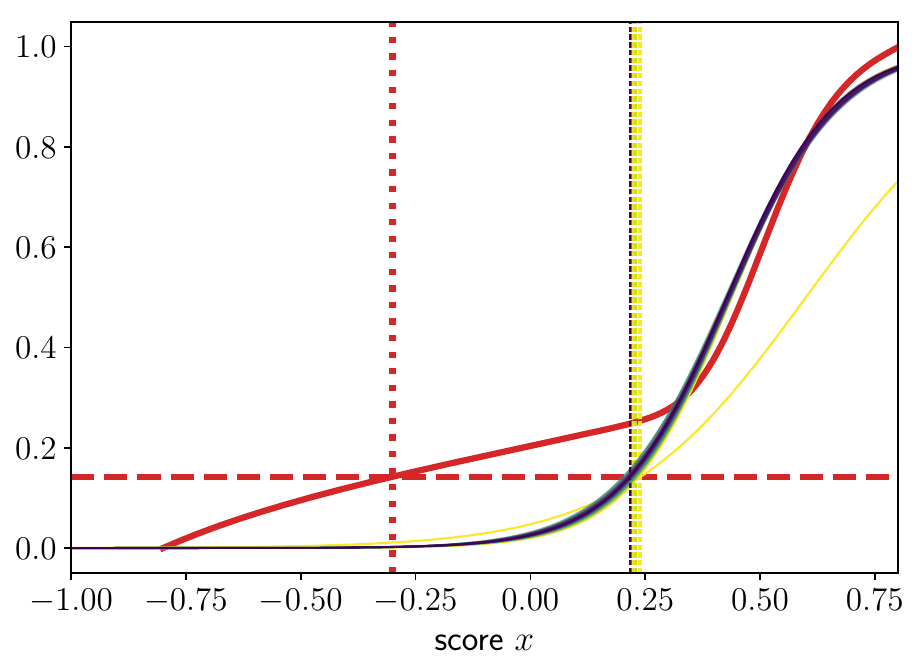}}
\hspace{0.5cm}
\subfloat[\tiny\textbf{deterministic, $t-1$}]{%
\includegraphics[width=\figwid \textwidth]{%
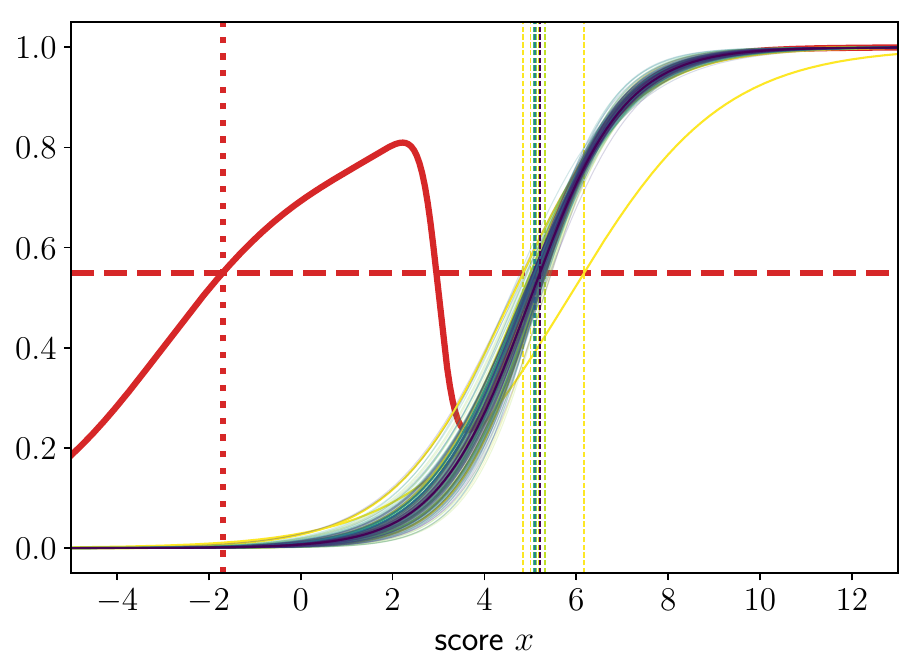}}
\\
\subfloat[\tiny\textbf{logistic, $t' < t$}]{%
\includegraphics[width=\figwid \textwidth]{%
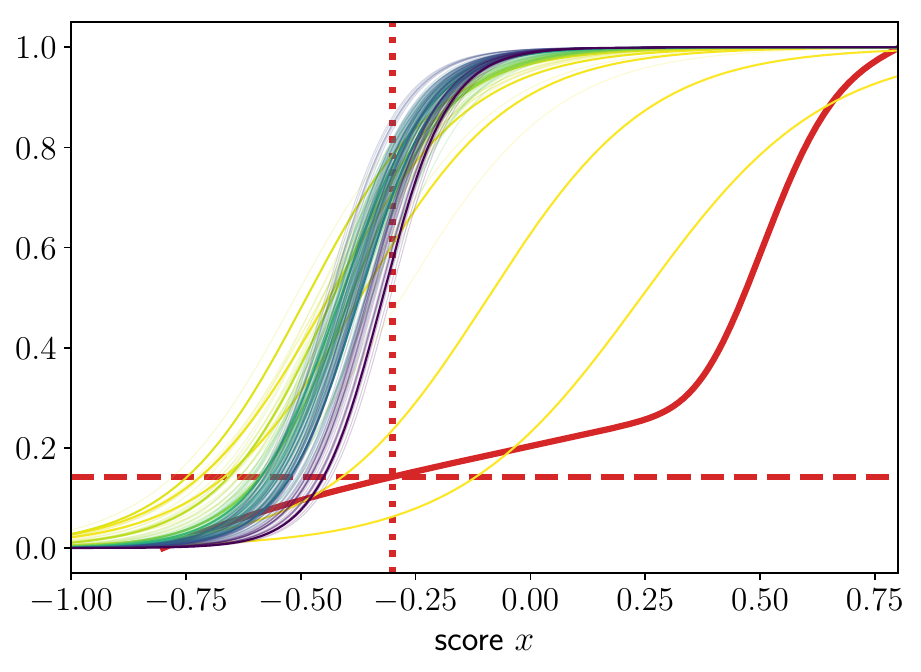}}
\hspace{0.5cm}
\subfloat[\tiny\textbf{logistic, $t' < t$}]{%
\includegraphics[width=\figwid \textwidth]{%
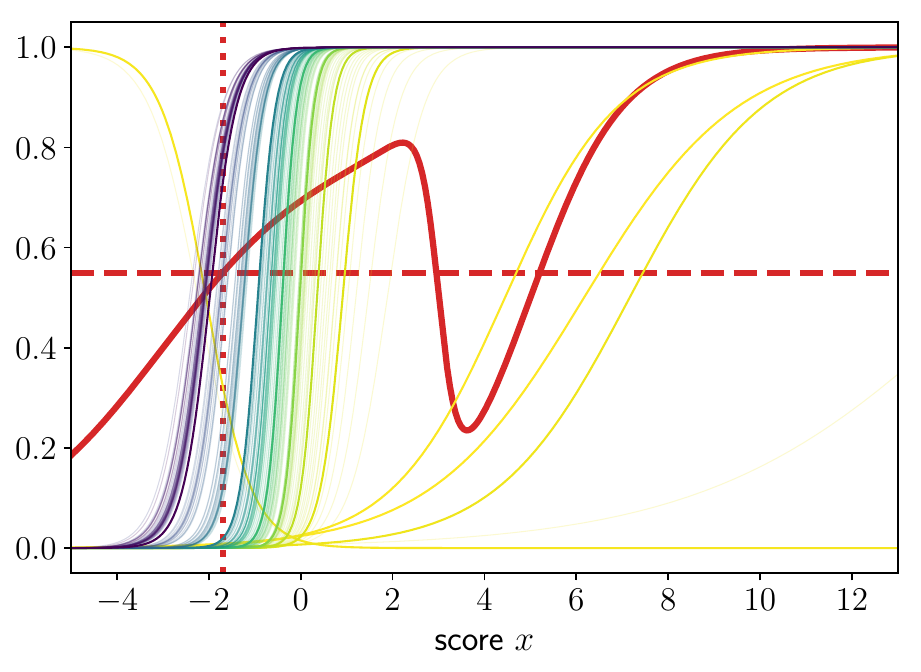}}
\\
\subfloat[\tiny\textbf{logistic, $t-1$}]{%
\includegraphics[width=\figwid \textwidth]{%
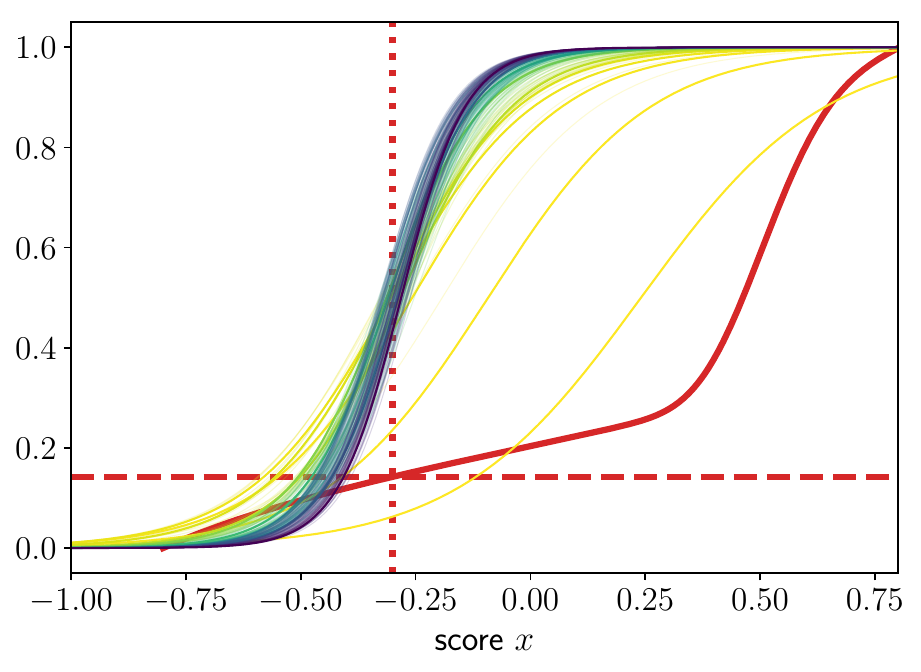}}
\hspace{0.5cm}
\subfloat[\tiny\textbf{logistic, $t-1$}]{%
\includegraphics[width=\figwid \textwidth]{%
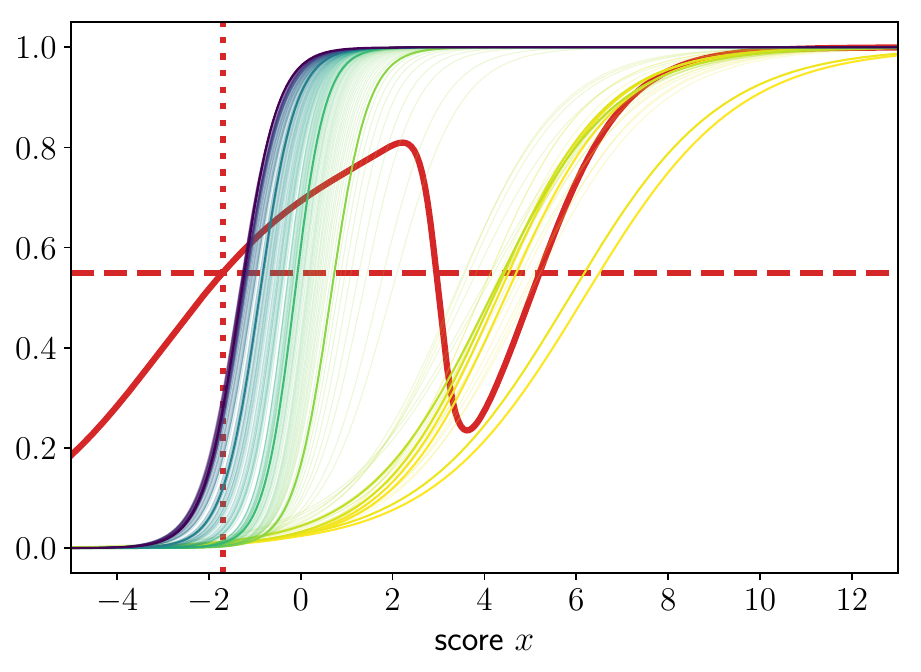}}
\\
\subfloat[\tiny\textbf{semi-logistic, $t'<t$}]{%
\includegraphics[width=\figwid \textwidth]{%
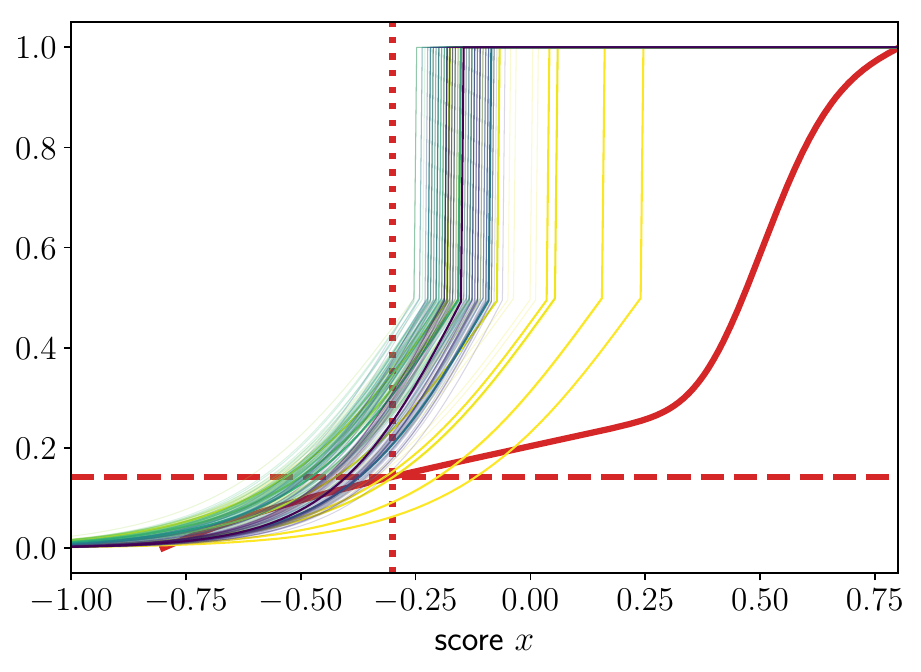}}
\hspace{0.5cm}
\subfloat[\tiny\textbf{semi-logistic, $t'<t$}]{%
\includegraphics[width=\figwid \textwidth]{%
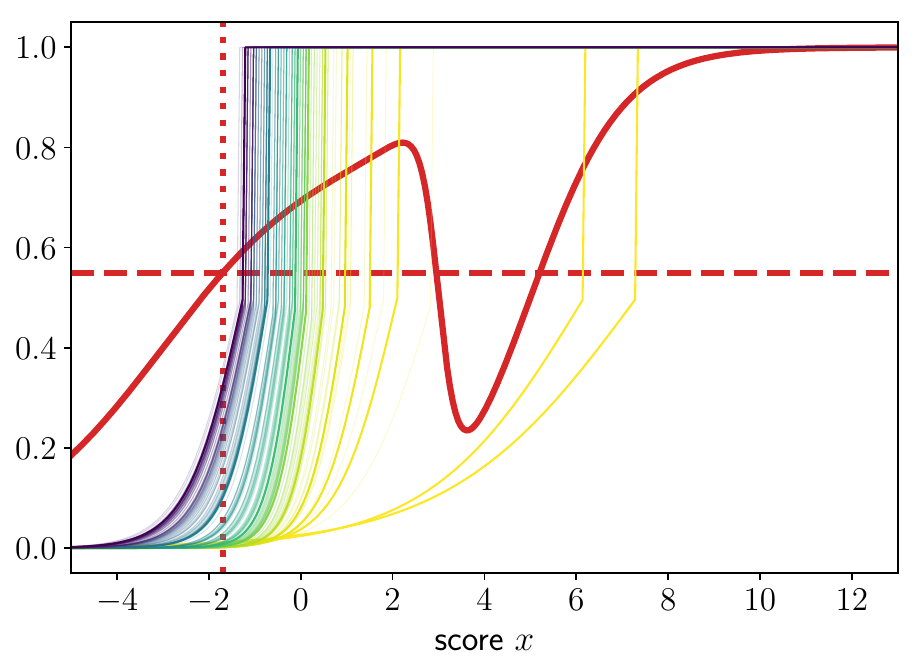}}
\\
\subfloat[\tiny\textbf{semi-logistic, $t-1$}]{%
\includegraphics[width=\figwid \textwidth]{%
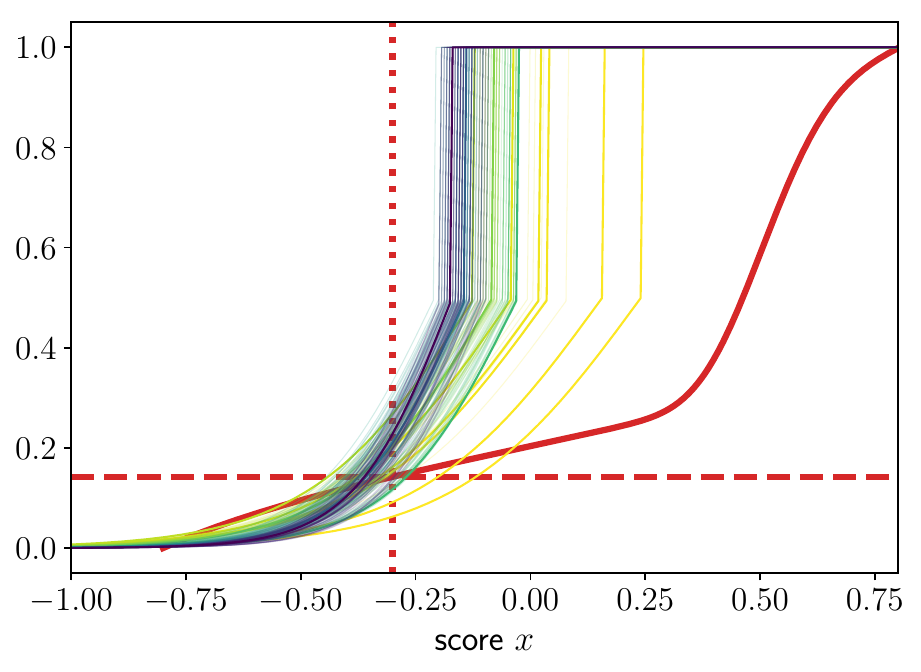}}
\hspace{0.5cm}
\subfloat[\tiny\textbf{semi-logistic, $t-1$}]{%
\includegraphics[width=\figwid \textwidth]{%
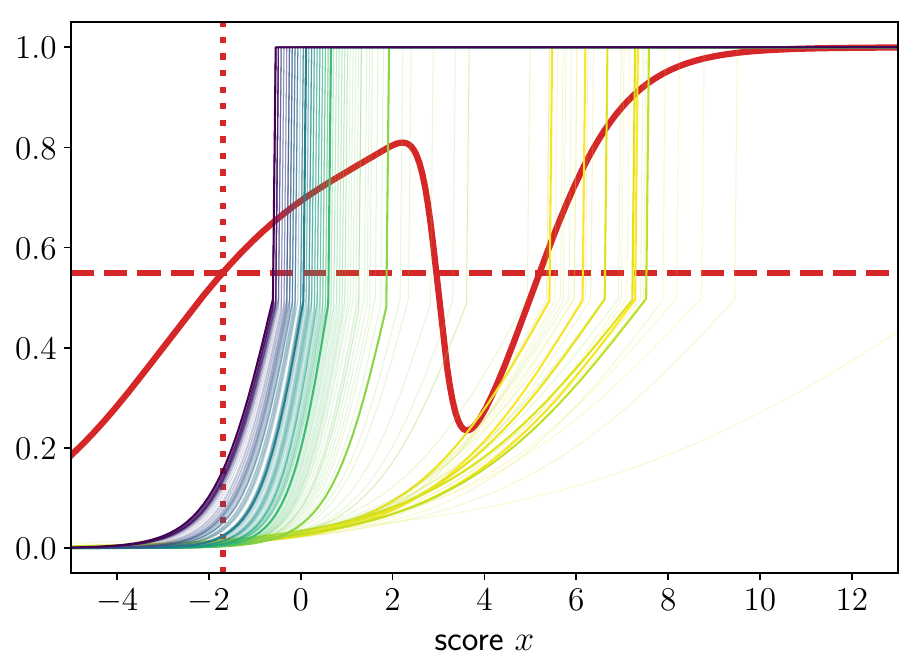}}
\caption{Learned predictive models for deterministic threshold rules and learned decision rules for the (semi-)logistic policies.
The columns correspond to the two synthetic settings.
We overlay the ground truth distribution $\dP(Y=1\given x)$ (red line), cost parameter $c$ (dashed, red), and optimal single decision boundary in $x$ within our model class (dotted, red).
We describe the plots in detail in the text.}
\label{fig:results_synthetic_evolution}
\end{figure}

\subsection{Experiments on synthetic data}

\xhdr{Setup}
The precise setup for the two different synthetic settings, illustrated in Figure~\ref{fig:synthetic-setting}, is as follows.
The only feature $x$ is a scalar score and $z \sim \mathrm{Ber}(0.5)$.
In the first setting, $x$ is sampled from a normal distribution $\mathcal{N}(\mu = 0.5 - z, \sigma = 1)$ truncated to $x \in [-0.8, 0.8]$, and the conditional probability $\dP(Y \given x)$ is strictly monotonic in the score and does not explicitly depend on $s$.
As a result, for any $c$, there exists a single decision boundary for the score that results in the optimal policy, which is contained in the class of logistic policies.
Note, however, that the score is not well calibrated, i.e., $\dP(Y \given x)$ is not directly proportional to~$x$.

In the second setting, $x \sim \mathcal{N}(\mu = 3 (0.5 - z), \sigma = 3.5)$.
Here, the conditional probability $\dP(Y \given x)$ crosses the cost threshold $c$ multiple times, resulting in two disjoint intervals of scores for which the optimal decision is $d=1$ (green areas).
Consequently, the optimal policy cannot be implemented by a deterministic threshold rule based on a logistic predictive model.
We show the best achievable single decision threshold in Figure~\ref{fig:synthetic-setting}.

\xhdr{Repeated figure}
First, in Figure~\ref{fig:results-synthetic-errs} we again show the contents of Figure~\ref{fig:results-synthetic} in the main text, but added effective utility and also show shaded regions for the 25th and 75th percentile over 30 runs.

\xhdr{Evolution of policies}
In Figure~\ref{fig:results_synthetic_evolution} we show for a representative run at $\lambda = 0$ how the different policies evolve in the two synthetic settings over time.
The two columns correspond to the two different synthetic settings.
For all policies, we show snapshots at a fixed number of logarithmically spaced time steps between $t=0$ and $t=200$.
For deterministic threshold rules, we show the logistic function of the underlying predictive model.
The vertical dashed line corresponds to the decision boundary in $x$.
For the logistic and semi-logistic policies, the lines correspond to $\pi_t(D=1 \given x)$, i.e., to the probability of giving a positive decision for a given input $x$.
Note that the semi-logistic policies have a discontinuity, because we do not randomize when the model believes $d = 1$ is a favorable decision with more than 50\% certainty.
For reference, we also show the true conditional distribution, the cost parameter, as well as the best achievable single decision boundary.

In the first setting, the exploring policies locate the optimal decision boundary, whereas the deterministic threshold rules, which are based on learned predictive models, do not, even though $\dP(Y = 1\given x)$ is monotonic in $x$ and has a sigmoidal shape.
The predictive models focus on fitting the rightmost part of the conditional well, but ignore the left region, from which they never receive data.

In the second setting, our methods explore more and eventually take positive decisions for $x$ right of the vertical dotted line in Figure~\ref{fig:synthetic-setting}, which is indeed the best achievable single threshold policy.
In contrast, non-exploring deterministic threshold rules again suffer from the same issue as in the first setting and converge to a suboptimal threshold at $x\approx 5$.
They ignore the left green region in Figure~\ref{fig:synthetic-setting} and do not overcome the dip of $\dP(Y=1\given x)$ below $c$, because they never receive data for $x \le 4$.

\begin{figure}
\centering
\def\figheight{3.5cm}
\includegraphics[width=\textwidth]{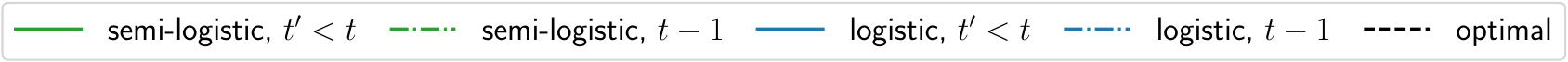}\\
\includegraphics[height=\figheight]{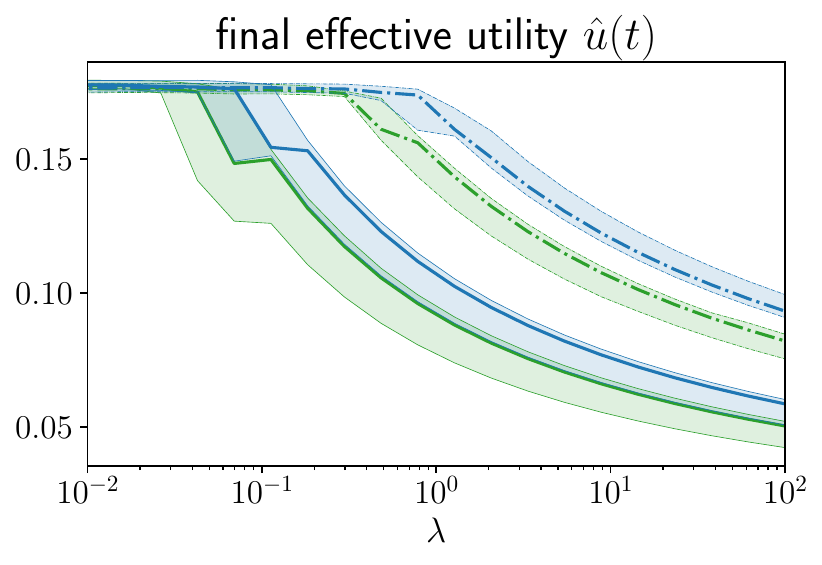}%
\hfill
\includegraphics[height=\figheight]{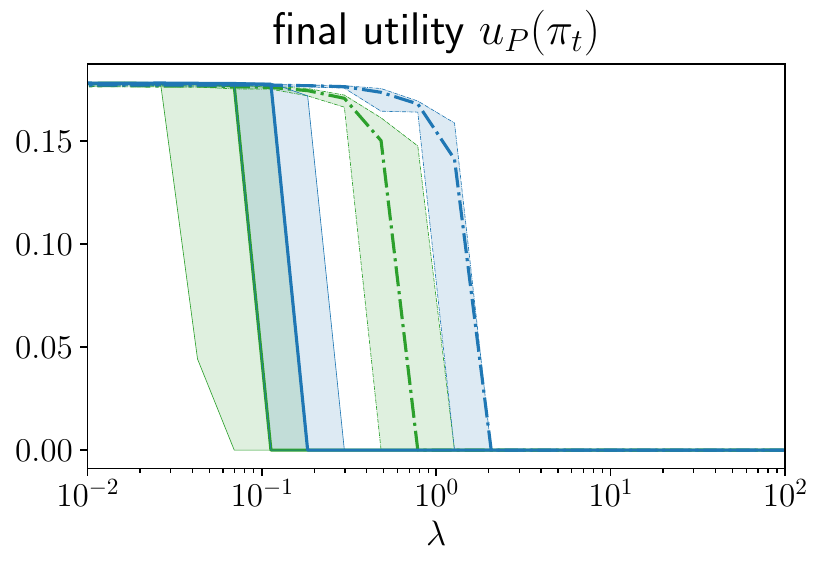}%
\hfill
\includegraphics[height=\figheight]{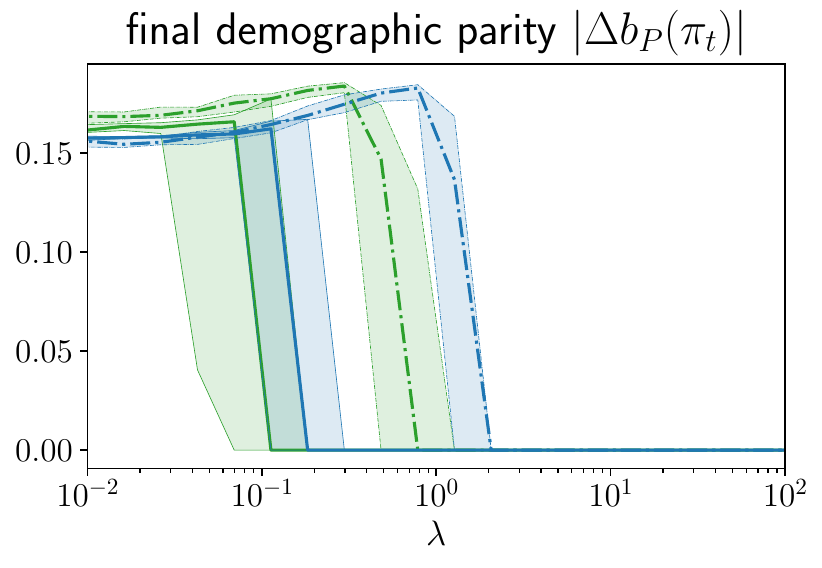}%
\\
\includegraphics[height=\figheight]{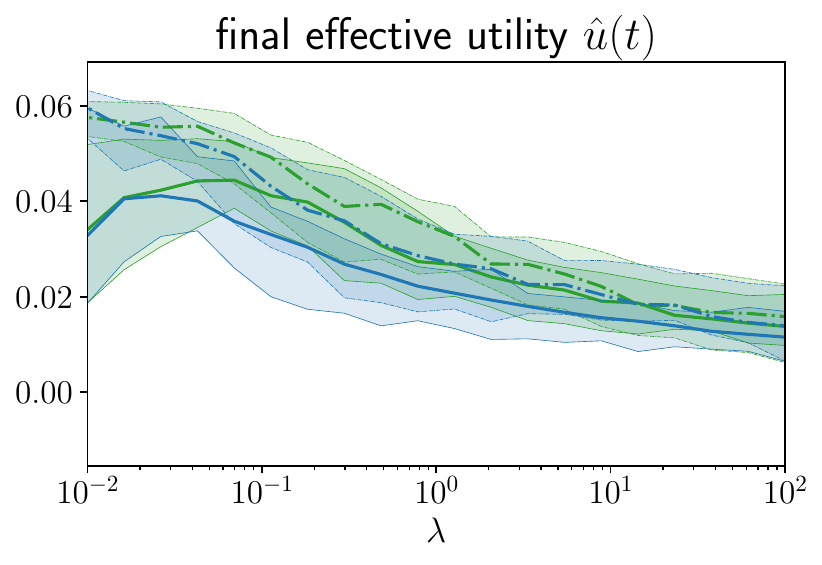}%
\hfill
\includegraphics[height=\figheight]{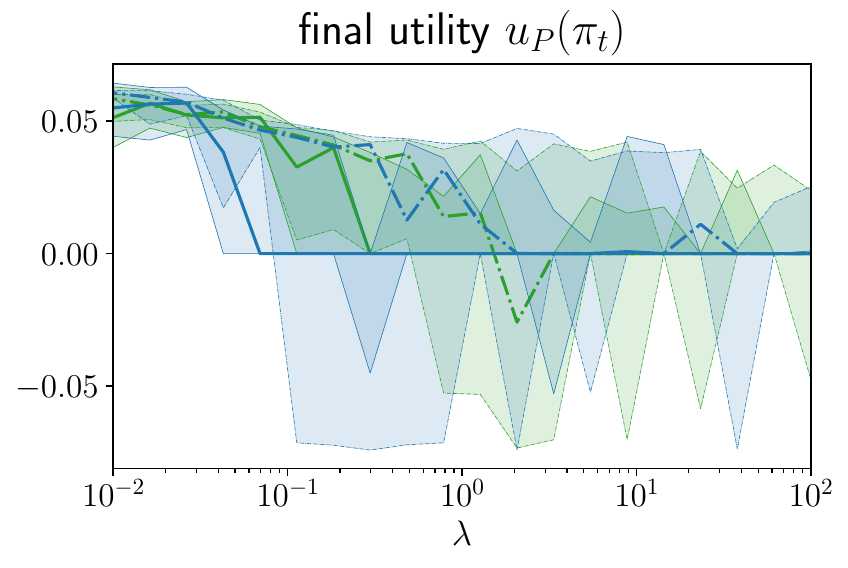}%
\hfill
\includegraphics[height=\figheight]{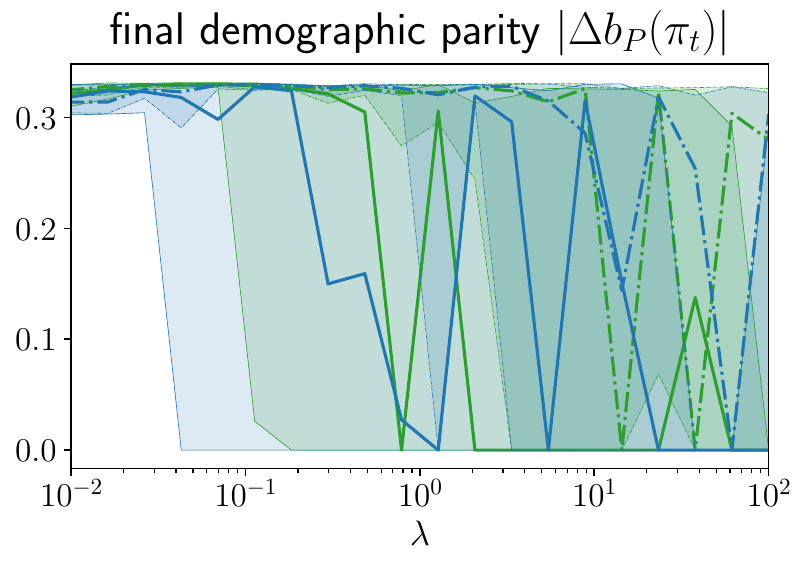}%
\caption{We show utility, effective utility, and demographic parity (columns) at the final time step $t=200$ as a function of $\lambda$ where we constrain demographic parity ($f(d, y) = d$).
The first row corresponds to the first setting and the second row corresponds to the second setting.}
\label{fig:results-synthetic-final-dp}
\end{figure}

\xhdr{Adding fairness constraints}
Figure~\ref{fig:results-synthetic-final-dp} shows how all metrics at the final time step $t = 200$ evolve as $\lambda$ is increased over the range $[10^{-0.5}, 10^4]$.
We use the benefit function for demographic parity in the fairness constraint, i.e., $f(d, y) = d$.
The first row corresponds to the first setting and the second row corresponds to the second setting.
In both cases, our approach achieves perfect fairness for sufficiently large $\lambda$ at the expected cost of a drop in (effective) utility.

\subsection{Experiments on real data}

\begin{figure}
\centering
\def\figheight{3.4cm}
\includegraphics[width=\textwidth]{legend-compas-time.pdf}\\
\includegraphics[height=\figheight]{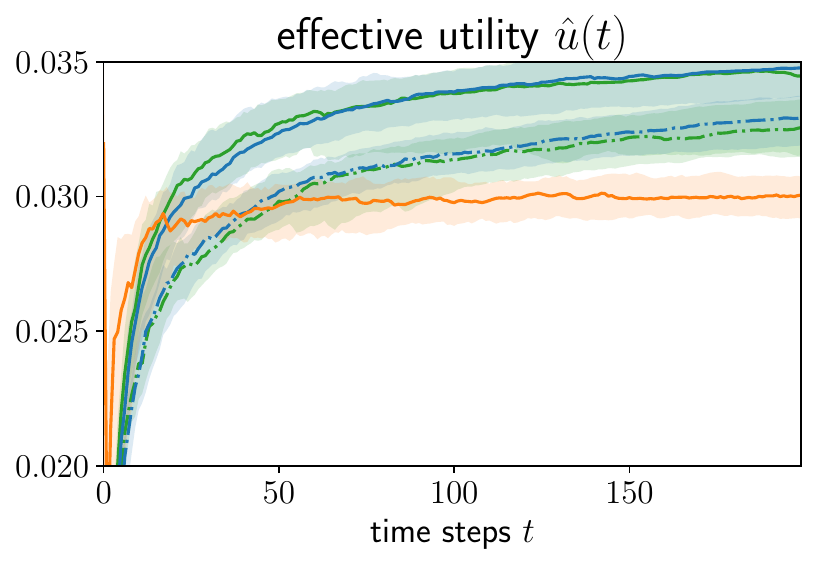}
\hfill
\includegraphics[height=\figheight]{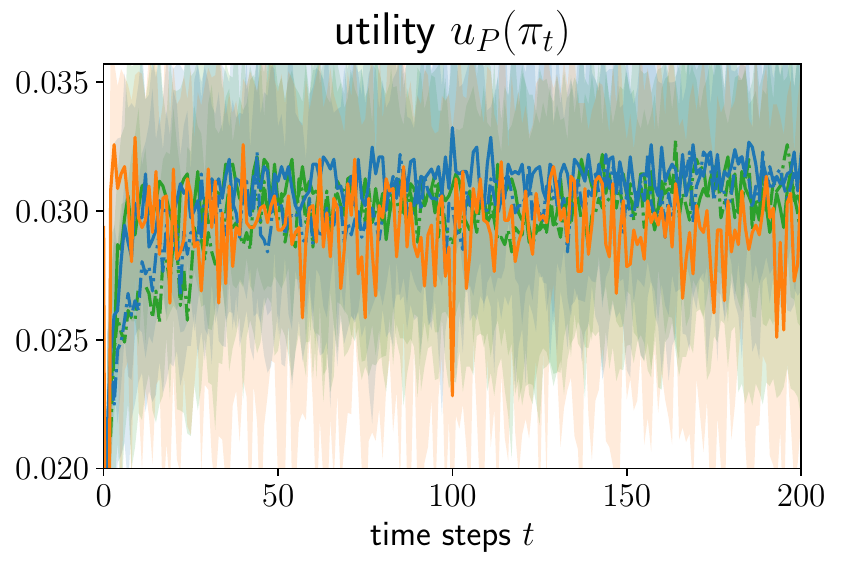}
\hfill
\includegraphics[height=\figheight]{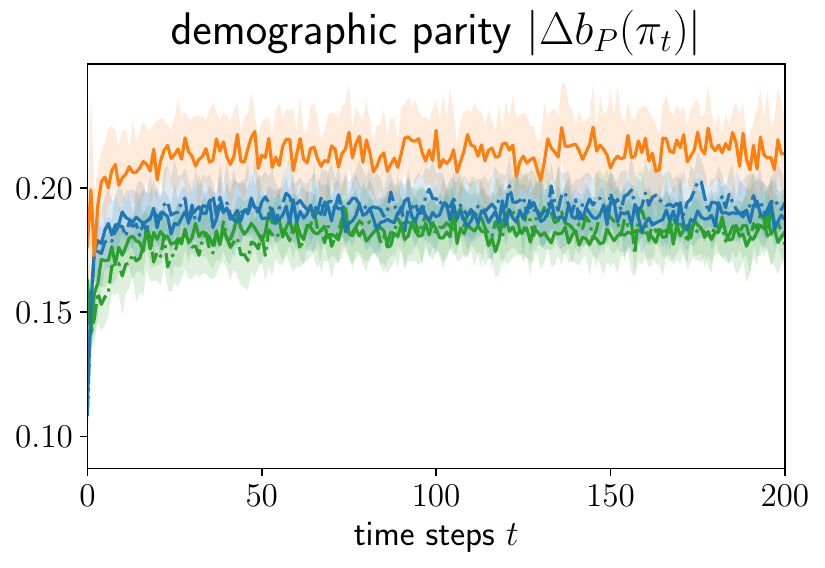}
\caption{Performance on COMPAS data.
We show the training progress for $\lambda = 0$, where all metrics are estimated on the held-out dataset.}
\label{fig:results-real-errs}
\end{figure}

\begin{figure}
\centering
\def\figheight{3.4cm}
\includegraphics[width=\textwidth]{legend-compas-lambda.pdf}\\
\includegraphics[height=\figheight]{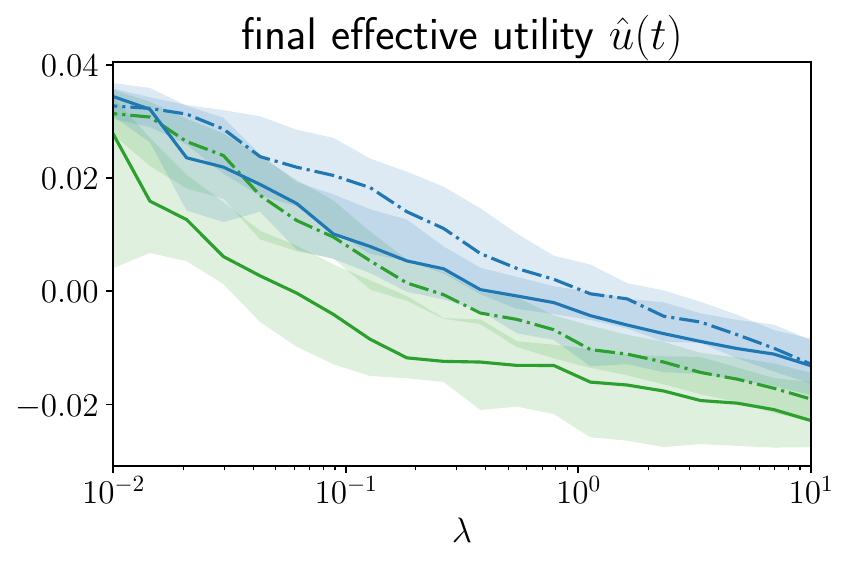}%
\hfill
\includegraphics[height=\figheight]{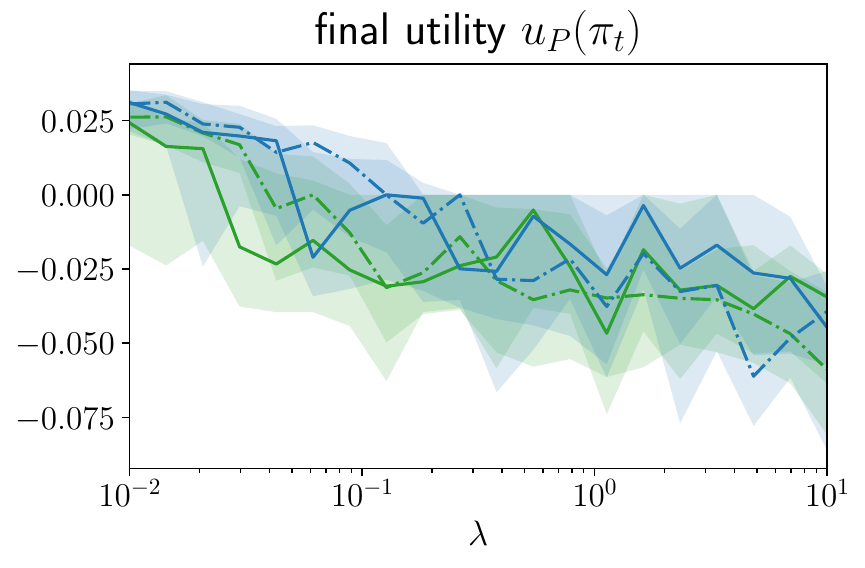}%
\hfill
\includegraphics[height=\figheight]{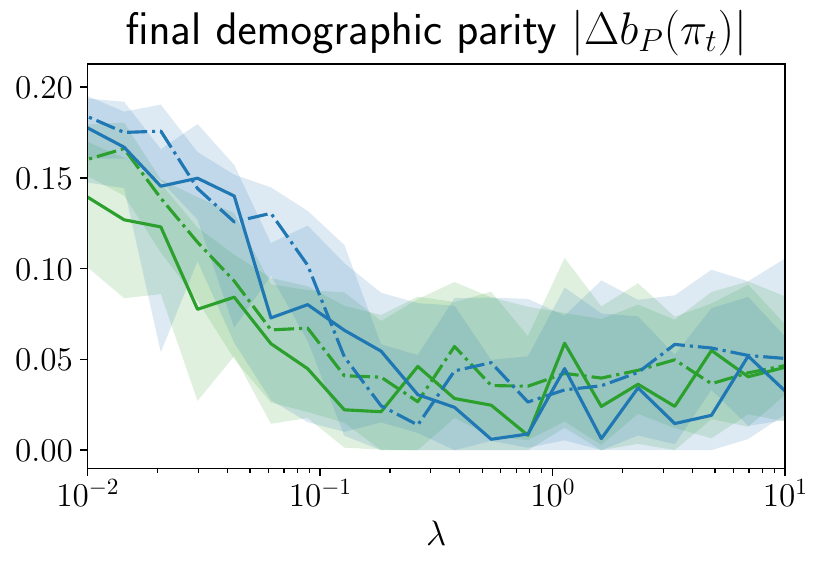}%
\caption{We show (effective) utility and demographic parity (columns) for the COMPAS dataset at the final time step $t=200$ estimated on the held-out dataset as a function of $\lambda$.}
\label{fig:results-real-final}
\end{figure}

First, in Figure~\ref{fig:results-real-errs} we again show the contents of Figure~\ref{fig:results-real-time} in the main text with shaded regions for the 25th and 75th percentile over 30 runs.
Analogously to Figure~\ref{fig:results-synthetic-final-dp}, we show the effect of enforcing fairness constraints in the COMPAS dataset in Figure~\ref{fig:results-real-final}.
The overall trends are similar to the results we have observed in the synthetic settings, reinforcing the applicability of our approach on real-world data.

\subsection{Parameter settings}
\label{sec:app:parameters}

The parameters used for the different experiments have been found by few iterations of manual trial.
The number of time steps is $T = 200$ for all datasets.
For the first synthetic setting we used $\alpha = 1$, $B = 256$, $M = 128$, $N = B \cdot M$, and $c \approx 0.142$ (chosen such that the optimal decision boundary is at $x = -0.3$).
For the second synthetic setting we used $\alpha = 0.5$, $B = 128$, $M = 32$, $N = B \cdot M$, and $c = 0.55$.
Here we also decay the learning rate by a factor of $0.8$ every 30 time steps.
For the COMPAS dataset we used $\alpha = 0.1$, $B = 64$, $M = 40\cdot B$, $N = B^2$, and $c = 0.6$.
While the initialization for the synthetic settings can be seen in Figure~\ref{fig:results_synthetic_evolution}, for COMPAS we trained a logistic predictive model on 500 i.i.d.\ examples for initializing policies and predictive models.

\end{document}